\documentclass[10pt,journal,compsoc]{IEEEtran}

\ifCLASSOPTIONcompsoc
  \usepackage[nospace,nocompress]{cite}
\else
  \usepackage[nospace,nocompress]{cite}
\fi

\usepackage{amsmath}
\usepackage{amssymb}
\usepackage{bm}
\usepackage{color}
\usepackage{multirow}
\usepackage{diagbox}
\usepackage{cancel}

\newcommand{\sgg}[1]{{\color{black}#1}}

\newcommand{\1}[1]{{\bf{\color{red}#1}}}

\newcommand{\3}[1]{{\bf{\color{blue}#1}}}
\usepackage[pagebackref=true,breaklinks=true,letterpaper=true,colorlinks,bookmarks=false]{hyperref}

\usepackage{pifont}
\newcommand{\cmark}{\ding{51}}%
\newcommand{\xmark}{\ding{55}}%

\ifCLASSINFOpdf
 \usepackage[pdftex]{graphicx}
  \graphicspath{{../pdf/}{../jpeg/}}
\else
\fi

\begin{document}
%
\title{Imbalanced Deep Learning by Minority Class Incremental Rectification}
%
%
%
%

\author{Qi Dong, Shaogang Gong, 
        and Xiatian Zhu
\IEEEcompsocitemizethanks{\IEEEcompsocthanksitem 
	Qi Dong and Shaogang Gong are with the School
of Electronic Engineering and Computer Science, Queen Mary University of London, UK. 
E-mail: \{q.dong, s.gong\}@qmul.ac.uk. 
Xiatian Zhu is with Vision Semantics Ltd., 
London, UK. 
E-mail: eddy@visionsemantics.com. }
}

\markboth{}%
{ Dong \MakeLowercase{\textit{et al.}}: 
	Imbalance Deep Learning.
}
%



\IEEEtitleabstractindextext{%
\begin{abstract}
Model learning from class imbalanced training data 
is a long-standing and significant challenge for machine learning. In particular,
existing deep learning methods consider mostly
either class balanced data or moderately imbalanced
data in model training, 
and ignore the challenge of learning from significantly imbalanced
training data.
To address this problem, we formulate a class imbalanced deep
learning model based on batch-wise incremental minority (sparsely
sampled) class rectification by hard sample mining in
majority (frequently sampled) classes during model training.
This model is designed to 
minimise the dominant effect of majority classes
by discovering sparsely sampled boundaries of minority classes in an
iterative batch-wise learning process.
To that end, we introduce a Class Rectification Loss (CRL)
function that can be deployed readily in deep network architectures. 
Extensive experimental evaluations are conducted
on three imbalanced person attribute benchmark datasets (CelebA,
X-Domain, DeepFashion) and one balanced object category benchmark
dataset (CIFAR-100). 
These experimental results demonstrate the performance advantages and
model scalability of the proposed batch-wise incremental minority
class rectification model over the existing 
state-of-the-art models for addressing the problem of imbalanced data learning. 

\end{abstract}

}

\maketitle

\IEEEdisplaynontitleabstractindextext

%
\IEEEpeerreviewmaketitle

\IEEEraisesectionheading{\section{Introduction}\label{sec:introduction}}


\IEEEPARstart{M}{achine} learning from class imbalanced data,
in which the distribution of training data across different object
classes is significantly skewed, 
is a long-standing problem
\cite{japkowicz2002class,weiss2004mining,he2009learning}.
%
Most existing learning algorithms 
produce
{\em inductive bias} (learning bias) towards the frequent
(majority) classes if training data are not balanced, resulting in
poor minority class recognition performance.
However, accurately detecting minority classes
is often important, e.g. in rare event discovery 
\cite{hospedales2013finding}.
%
%
%
A simple approach to overcoming class imbalance in model learning is to re-sample the 
training data (a pre-process), e.g. by down-sampling majority classes,
over-sampling minority classes, or some combinations
\cite{drummond2003c4,chawla2002smote,maciejewski2011local}.
%
Another common approach is cost-sensitive learning, 
which reformulates existing learning algorithms by weighting the
minority classes more 
\cite{ting2000comparative,tang2009svms,Akbani-ecml04}.

%

\begin{figure} [t]
	\includegraphics[width=1\linewidth]{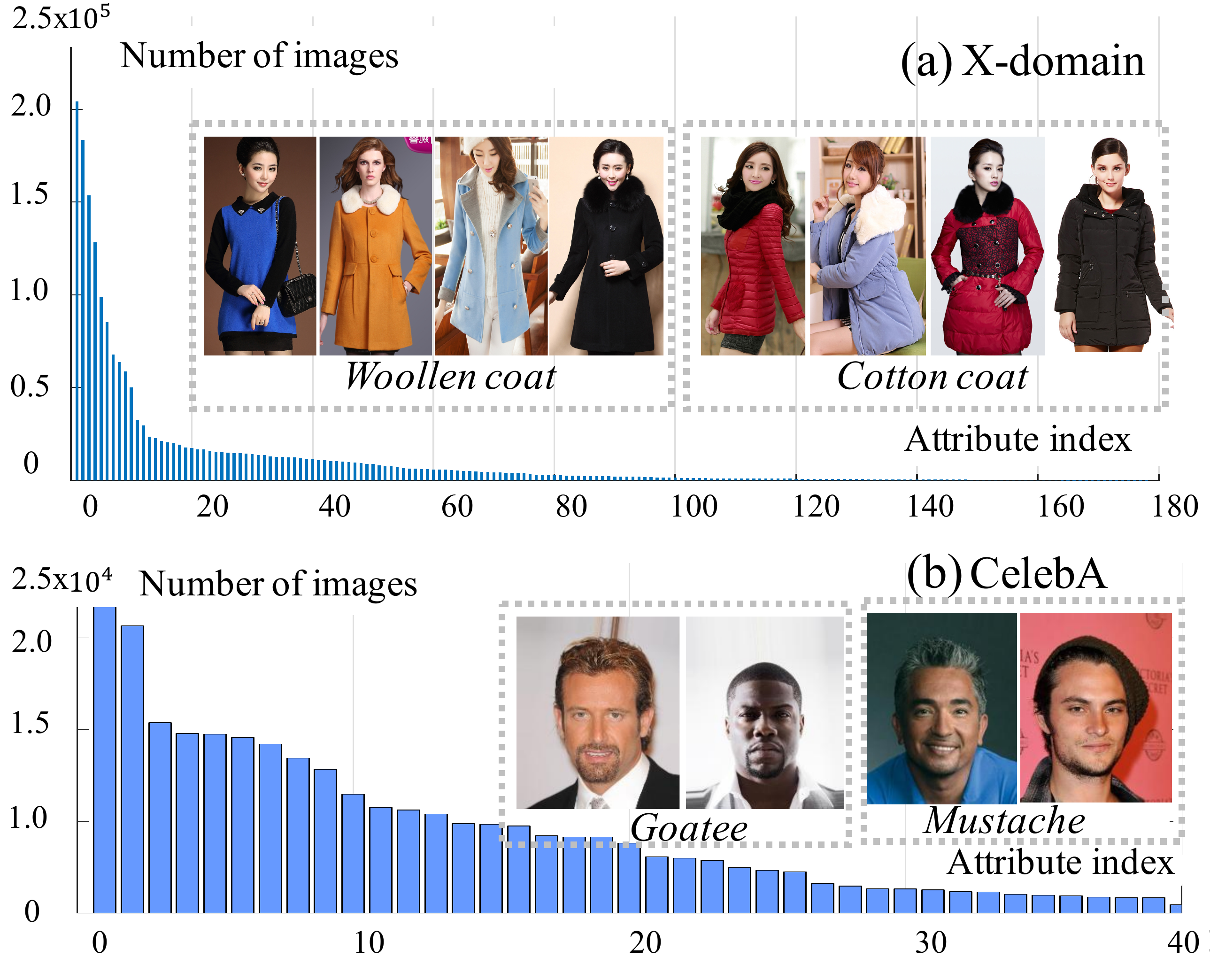}
	\vskip -0.2cm
	\caption{Imbalanced training data class distributions: (a) clothing
		attributes (X-Domain \cite{chen2015deep}), (b) facial attributes 
		(CelebA \cite{liu2015deep}).}
	\label{fig:problem}
	\vspace{-0.2cm}
\end{figure}

Over the past two decades, a range of class imbalanced learning methods have been developed \cite{krawczyk2016learning}.
However, they mainly investigate the single-label binary-class
imbalanced learning problem
in small scale data with 
class imbalance ratios being small, e.g. within 1:100.
These methods are limited when applied to learning from big scale data in
computer vision.
Visual data are often interpreted by multi-label semantics, 
e.g. web person images with multi-attributes on clothing and facial
characteristics. Automatic recognition of these nameable properties is
very useful 
\cite{gong2014person,feris2014attribute}, but challenging for model
learning due to: 
(1) Very large scale imbalanced training data 
\cite{Akbani-ecml04,ChenEtAlcvpr13,huang2016learning},
with clothing and facial attribute labelled data exhibiting
power-law distributions (Fig. \ref{fig:problem}).  
(2) Subtle appearance discrepancy between different fine-grained
attribute classes, e.g. 
``Woollen-Coat'' can appear very similar to ``Cotton-Coat'', 
whilst ``Mustache" can be visually hard to be distinct
(Fig. \ref{fig:problem}(b)). 
%
To discriminate subtle classes 
from multi-labelled
images at large scale,
standard learning algorithms require a vast 
quantity of class {\em balanced} training data for all labels
\cite{chen2015deep,zhang2014panda}. 

There is a vast quantity of severely imbalanced visual data on the
Internet. Conventional learning algorithms
\cite{triguero2015rosefw,krawczyk2016learning} are poorly suited for three
reasons: {\em First}, conventional imbalanced data learning
methods without deep learning rely on 
hand-crafted features extracted from small data, 
which are inferior to big data deep learning
based richer feature representations 
\cite{simonyan2014very,sharif2014cnn,krizhevsky2012imagenet,bengio2013representation}.
{\em Second}, 
deep learning in itself also suffers 
from class imbalanced training data 
\cite{zhou2006training,jeatrakul2010classification,huang2016learning}
(Table \ref{tab:ration} and Sec. \ref{sec:exp_cifar100}).
{\em Third}, directly incorporating
existing imbalanced data learning algorithms into a deep learning framework
does not provide effective solutions
\cite{alejo2006improving,khoshgoftaar2010supervised,mazurowski2008training}.

Overall, imbalanced big data deep learning is under-studied partly due
to that popular image benchmarks for large scale deep
learning, e.g. ILSVRC, do not exhibit significant class imbalance after
some careful sample filtering being applied in 
those benchmark constructions (Table \ref{tab:dataset_imbalance}). 
More recently, there are a few emerging large scale clothing and
facial attribute datasets that are significantly more
imbalanced in class labelled data distributions (Fig.~\ref{fig:problem}),
as these datasets are drawn from online Internet sources without
artificial sample filtering
\cite{chen2015deep,huang2015cross,liu2016deepfashion,liu2015deep}. 
For example,
the {\em imbalance-ratio} (lower is more extreme) 
between the minority
classes 
and the majority classes 
in the CelebA face attribute dataset \cite{liu2015deep}
is 
1:43 (3,713 : 159,057 samples), whilst the X-Domain clothing attributes
are even more imbalanced with an imbalance-ratio of 1:4,162 (20 : 204,177)
\cite{chen2015deep} (Table \ref{tab:dataset_imbalance}).

%

\begin{table} 
	\scriptsize
	\centering
	\setlength{\tabcolsep}{0.15cm}
	\caption{
		Comparing large datasets w.r.t. training data imbalance in terms of 
		class {\em imbalance ratio} (the sample size ratio
                between the smallest and largest classes).
		The ratios are for the standard
		train/val/test data split if available, otherwise the whole
		dataset. 
	For MS-COCO \cite{lin2014microsoft}, no numbers are available
	for calculating the imbalance ratio, because their
	images often contain simultaneously multiple 
	classes of objects 
	and multiple instances of a specific class.}
	\vskip -0.3cm
	\label{tab:dataset_imbalance}
	\begin{tabular}{c|c|c|c}
		\hline
		ILSVRC2012-14 \cite{russakovsky2015imagenet} 
		%
		& MS-COCO \cite{lin2014microsoft} 
		& VOC2012 \cite{everingham2015pascal}
		& CIFAR-100 \cite{krizhevsky2009learning} \\
		\hline
		1 : 2 
		& - & 1 : 13 
		& 1 : 1 \\ \hline \hline 
		Caltech 256 \cite{griffin2007caltech} 
		& CelebA \cite{liu2015deep} 
		& DeepFashion \cite{liu2016deepfashion}
		& X-Domain \cite{chen2015deep}  
		\\ \hline
		1 : 1 & 1 : 43 
		&  1 : 733 
		& \bf 1 : 4,162 
		\\ 
		\hline 
	\end{tabular}
	\vspace{-0.5cm}
\end{table}

\begin{figure*}[t]
	\centering
	\includegraphics[width=0.99\linewidth]{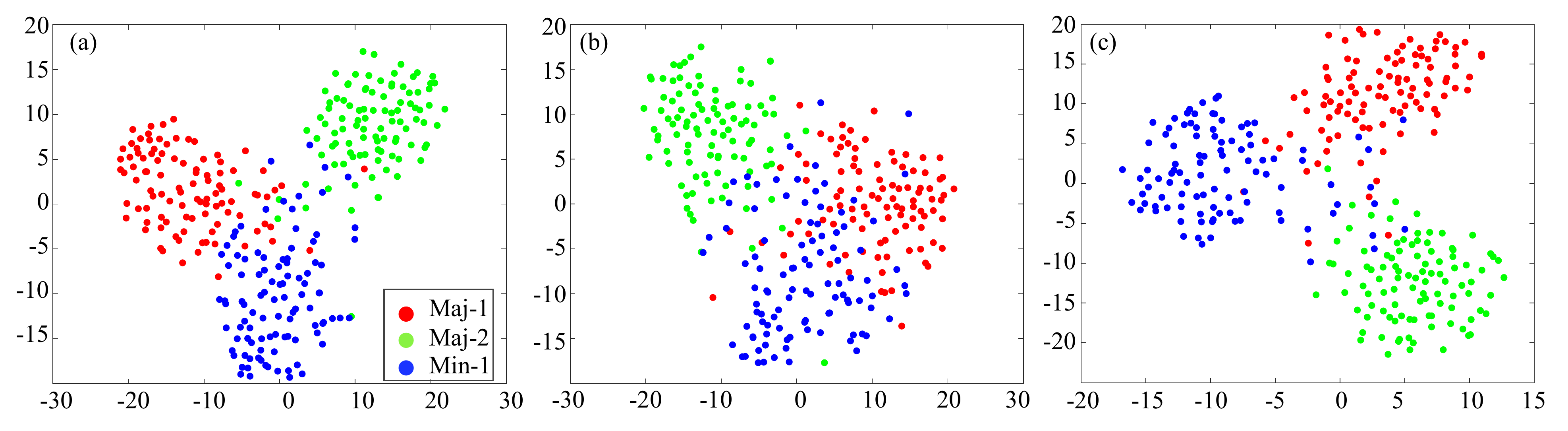}
	\vskip -0.3cm
	\caption{Visualisation of deep feature distributions learned by the ResNet32 \cite{he2016deep} models  
		trained on 
		{\bf (a)} class balanced data,
		{\bf (b)} class imbalanced data, and
		{\bf (c)} class imbalanced data with our
		CRL model. Customised CIFAR-100 object category training image sets were
		used 
		(more details in Sec. \ref{sec:exp_cifar100}).
		In the illustration,
		we showed only 3 (2 majority and 1 minority) classes
		for the visual clarity sake.
		It is observed that the minority class (Min-1)  
		is overwhelmed more severely by the two majority classes (Maj-1, Maj-2)
		when deploying the existing CNN model
		given class imbalanced training data. 
		The proposed CRL approach rectifies clearly this undesired 
		model induction bias 
		inherent to existing deep learning architectures.
	}
	\label{fig:feat_visualization}
	\vspace{-0.3cm}
\end{figure*}

This work addresses the problem of large scale imbalanced 
data deep learning for multi-label classification.
This problem is characterised by 
{(1)} Large scale training data; 
{(2)} Multi-label per data sample;
{(3)} Extremely imbalanced training data with an imbalance-ratio
being greater than 1:1000;
{(4)} Variable per-label attribute values, ranging from binary to
multiple attribute values per label.
%
The {\bf contributions} of this work are:
{\bf (I)} We solve the {\em large scale imbalanced data deep learning} problem. 
This differs from the conventional imbalanced data learning studies
focusing on small scale data single-labelled with a limited number of  
classes and small data imbalance-ratio.
%
{\bf (II)} We present a novel approach to imbalanced deep learning by
minority class incremental rectification using {\em batch-wise mining}
of hard samples 
on the minority
classes in a {\em batch-wise optimisation} process.
This differs from contemporary multi-label learning methods
\cite{chen2015deep,huang2015cross,liu2016deepfashion,dong2016multi,zhang2014panda}, which either assume class balanced training data or simply ignore the
imbalanced data learning problem all together. 
%
%
{\bf (III)} 
We formulate a {\em Class Rectification Loss} (CRL)
regularisation algorithm for minority class
incremental rectification.
In particular, the computational complexity of imposing this
rectification loss is restrained by iterative mini-batch-wise model
optimisation (small data pools). This is in contrast to the global
model optimisation over the entire training data pool of the Large 
Margin Local Embedding (LMLE) algorithm\footnote{ 
In LMLE, a computationally expensive data pre-processing 
(including clustering and quintuplet construction)
is required for each round of deep model learning. In particular,
to cluster $n$ (e.g. 150,000+)
training images w.r.t. an attribute label by
$k$-means, its operation complexity is super-polynomial
with the need for at least $2^{\Omega(\sqrt{n})}$ 
($\Omega$ the lower bound complexity) iterations of
cluster refinement on $n$ samples \cite{arthur2006slow}. 
As each iteration is linear to $k$ and $n$,
a clustering takes the complexity at $k \times O(n) \times 2^{\Omega(\sqrt{n})}$
($O$ the upper bound complexity).
To create a quintuplet for each data sample,
four cluster- and class-level searches are needed, each proportion to
the training data size $n$ with the overall search 
complexity as quadratic to $n$ ($O(n^2)$).
Given a large scale training set, 
it is likely that this pairwise search part takes
the most significant cost in the pre-processing. 
Both clustering and quintuplet operations are needed for each
attribute label, and their costs are proportional to 
the total number $n_\text{val}$ of attribute values, e.g. 80 times for CelebA and 178 times for
X-domain. 
Consequently, the total complexity of the pre-processing per round
is $n_\text{val} \times \big( k \times O(n) \times 2^{\Omega(\sqrt{n})} + O(n^2)\big)$.}
\cite{huang2016learning}.
%
There are two advantages of our approach: {\em First}, the model only
requires incremental class imbalanced data learning 
for all attribute labels concurrently without any 
additional single-label sampling assumption 
(e.g. per-label oriented quintuplet construction);
{\em Second}, model learning is independent to the overall training
data size, the number of class labels, and without pre-determined
global data clustering. This makes the model much more scalable to
learning from large training data. 
%

%
%
%
%
Extensive evaluations were performed on the CelebA face attribute
\cite{liu2015deep} and X-Domain clothing attribute \cite{chen2015deep}
benchmarks, with further evaluation on the DeepFashion
clothing attribute \cite{liu2016deepfashion} benchmark.
These experimental results show a clear advantage of
CRL over 12 state-of-the-art models compared, including 7
attribute models
(PANDA \cite{zhang2014panda},
ANet \cite{liu2015deep},
Triplet-$k$NN \cite{schroff2015facenet},
FashionNet \cite{liu2016deepfashion}, 
DARN \cite{huang2015cross}, 
LMLE \cite{huang2016learning}, 
MTCT \cite{dong2016multi}).
We further evaluated the CRL method 
on the class balanced single-label object recognition benchmark
CIFAR-100 \cite{krizhevsky2009learning},
and constructed different class imbalance-ratios therein to 
quantify the
model performance gains under controlled varying
degrees of imbalance-ratio in training data.

\section{Related work}\label{sec:relatedworks}

\noindent {\bf Class imbalanced learning} 
aims to mitigate model learning bias towards majority classes
by lifting the importance of minority classes
\cite{japkowicz2002class,weiss2004mining,he2009learning}.
Existing methods include:
(1) \textbf{\em Data-level}:
Aiming to rebalance the class prior distributions in a pre-processing procedure.
This scheme is attractive as the only change needed is to the training data 
rather than to the learning algorithms.
Typical methods include down-sampling majority classes, over-sampling minority classes,
or both \cite{chawla2002smote,drummond2003c4,han2005borderline,
	he2009learning,maciejewski2011local,oquab2014learning}.
However, over-sampling can easily cause model overfitting owing to
repeatedly visiting duplicated samples \cite{chawla2002smote}. 
Down-sampling, on the other hand, throws away valuable information
\cite{drummond2003c4,japkowicz2000learning}.
(2) \textbf{\em Algorithm-level}:
Modifying existing algorithms to give more emphasis on the
minority classes \cite{barandela2003strategies,chen2004using,Akbani-ecml04,lin2002support,liu2000improving,zadrozny2001learning,quinlan1991improved,wu2005kba,tang2009svms,zadrozny2003cost,ting2000comparative}.
One strategy is the cost-sensitive learning which
assigns varying costs to different classes,
e.g. a higher penalty for minority class samples 
%
\cite{weiss2004mining,ting2000comparative,zadrozny2003cost,chen2004using,
	zhou2006training,tang2009svms}.
However, it is in general difficult to optimise the cost matrix or
relationships. Often, it is given by experts therefore
problem-specific and non-scalable.
In contrast, the threshold-adjustment technique changes the decision threshold 
in test time
\cite{zhou2006training,provost2000machine,krawczyk2015cost,yu2016odoc,chen2006decision}.
(3) \textbf{\em Hybrid}: 
Combined data-level and algorithm-level
rebalancing \cite{wozniak2013hybrid,wang2012applying,wozniak2014survey}.
These methods still only consider
small scale imbalanced data learning, characterised by:
(a) Limited number of data samples and classes, 
(b) Non-extreme imbalance ratios,
(c) Single-label classification,
(d) Problem-specific hand-crafted low-dimensional features. 
%
In large, these classical techniques are poor for severely
imbalanced data learning given big visual data.
%
%

There are early neural network based methods for imbalanced data learning
\cite{zhou2006training,jeatrakul2010classification,lan2010investigation,huang2006evaluation,fernandez2011dynamic,castro2013novel,alejo2006improving,khoshgoftaar2010supervised,mazurowski2008training}.  
However, these works still only address small scale imbalanced
data learning with neural networks merely acting as nonlinear
classifiers without end-to-end learning. 
%
A few recent studies
\cite{khan2017cost,oquab2014learning,shen2015deepcontour,wang2016training,guan2015deep}
have exploited classic strategies 
in single-label deep learning. 
For example, 
the binary-class classification problem is studied by
per-class mean square error loss \cite{wang2016training},
synthetic minority sample selection \cite{guan2015deep},
and constant class ratio in mini-batch \cite{yan2015deep}.
The multi-class classification problem is addressed by online
cost-sensitive loss \cite{khan2017cost}. 
More recently,
the idea of preserving local class structures (LMLE) was proposed for
imbalanced data deep learning, 
but without end-to-end model training and with only single-label
oriented training unit design \cite{huang2016learning}.  
In contrast,
our model is designed for end-to-end imbalanced data deep learning
for multi-label classification, scalable to large training data. 

\noindent {\bf Hard sample mining} 
has been extensively exploited in computer vision, 
e.g. object detection \cite{felzenszwalb2010object,shrivastava2016training},
face recognition \cite{schroff2015facenet},
image categorisation \cite{oh2016deep}, 
and unsupervised representation learning \cite{wang2015unsupervised}. 
The rational for mining {\em hard}
negatives (unexpected) is that, they are more informative than {\em easy} negatives (expected)
as they violate a model class boundary 
by being on
the wrong side and also far away from it. Therefore, hard negative
mining enables a model to improve itself quicker and more
effectively with less training data. Similarly, model learning can also benefit
from mining hard positives (unexpected), i.e. those on the correct
side but very close to or even across 
a model class boundary. 
In our model learning, we {\em only} consider hard mining on
the minority classes for efficiency. 
Moreover, our batch-balancing hard mining strategy 
differs from that of LMLE \cite{huang2016learning}
by eliminating exhaustive searching of the entire training
set (all classes), hence computationally more scalable than LMLE.

\noindent {\bf Deep metric learning}
is based on the idea of combining deep neural networks
with metric loss functions in a joint end-to-end learning process
\cite{Ustinova2016hist,schroff2015facenet,wang2014learning,oh2016deep}. 
Whilst adopting similarly a generic margin based 
loss function \cite{liu2009learning,chopra2005learning}, 
deep metric learning does not consider the class imbalanced data learning
problem. In contrast, our method is specifically designed
to address this problem by incrementally 
rectifying the structural significance of minority classes in a
batch-wise end-to-end learning process, so to achieve scalable
imbalanced data deep learning. 

\noindent {\bf Deep learning of clothing and facial attributes}
%
has been recently exploited 
\cite{chen2015deep,huang2015cross,liu2016deepfashion,dong2016multi,liu2015deep,zhang2014panda},
given the availability of large scale datasets and deep models'
strong capacity for learning from big training data. 
However, existing methods ignore mostly imbalanced
class data distributions, 
resulting in suboptimal model learning and poor model performance on
the minority classes. 
One exception is the LMLE model \cite{huang2016learning} which
studies imbalanced data deep learning \cite{he2009learning}.
Compared to our end-to-end learning using mini-batch hard sample
mining on the minority classes only, 
LMLE is not end-to-end learning and with global hard mining over the
entire training data, it is computationally complex and expensive,
not lending itself naturally to big training data. 
\section{Scalable Imbalanced Deep Learning} \label{method}

\begin{figure}[th]
	\centering
	\includegraphics[width=1\linewidth]{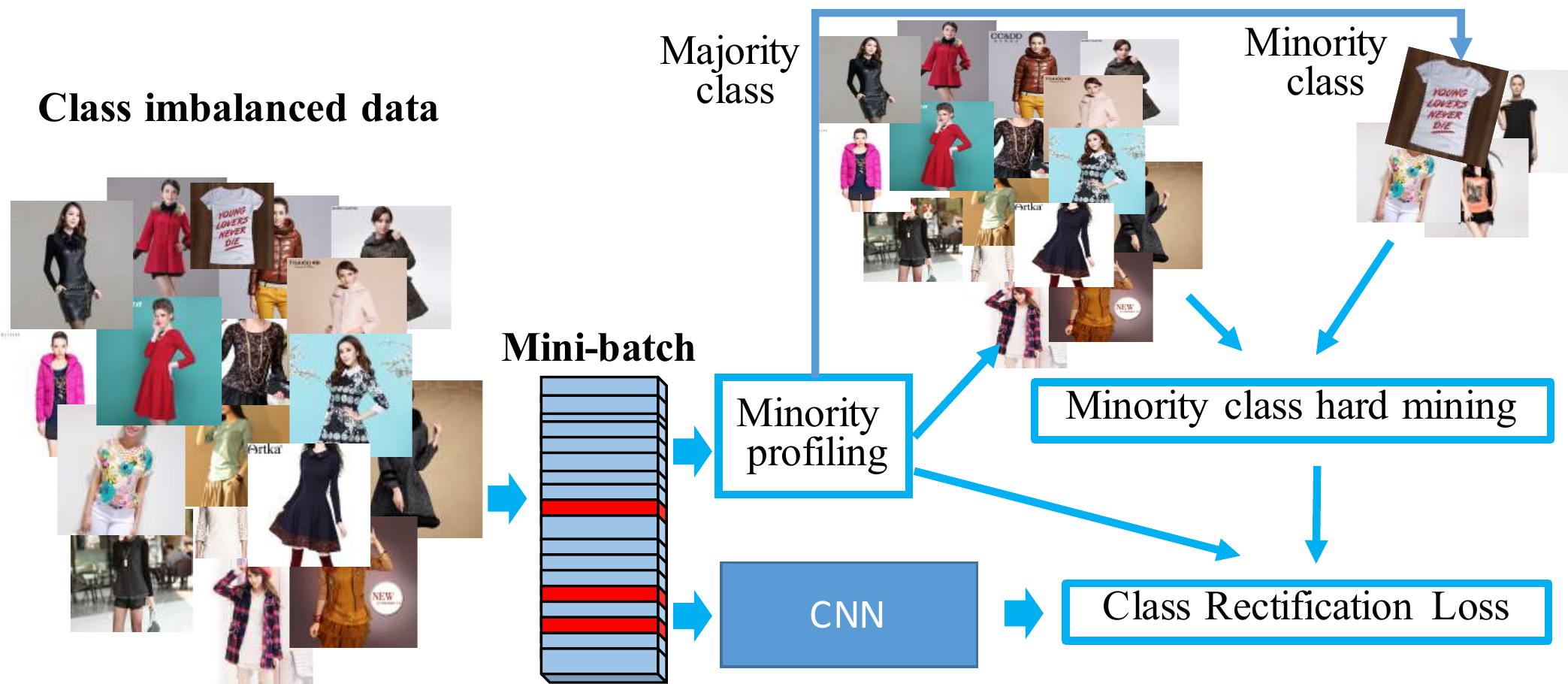}
	\vskip -0.3cm
	\caption{
		Overview of the proposed Class Rectification Loss (CRL) regularisation
		for large scale class imbalanced deep learning.
	}
	\vspace{-0.3cm}
	\label{fig:pipeline}
\end{figure}

For the problem of imbalanced data deep
learning from large training data, we consider the
problem of person attribute recognition, both facial and clothing attributes.
This is a multi-label multi-class learning problem given imbalanced
training data, a generalisation of the more common single-label
binary-/multi-class recognition problem. 
Specifically, we wish to construct a deep learning model capable of recognising {\em multi-labelled} person attributes
$\{z_j\}_{j=1}^{n_\text{attr}}$ in web
images, with a total of $n_\text{attr}$ different attribute labels.
Each {\em label} $z_j$ has its respective {\em class} value range $Z_j$,
e.g. multi-class 
clothing attribute or 
binary-class 
facial attribute. 
%
%
Suppose we have 
$n$ training images
$\{\bm{I}_i\}_{i=1}^{n}$ with their attribute annotation vectors
$\{\bm{a}_i\}_{i=1}^{n}$,
and $\bm{a}_i=[a_{i,1}, \dots, a_{i,j}, \dots,a_{i,n_\text{attr}}]$
where $a_{i,j}$ refers to the $j$-th attribute class value of the image $\bm{I}_i$.
The number of images available for different attribute classes
varies greatly (Fig.~\ref{fig:problem}) therefore imposing a
significant {\em multi-label imbalanced class data} distribution challenge
to model learning.
Most attributes are {\em localised} to image regions, even though the
location information is not annotated ({\em weakly labelled}).
%
%
We consider to jointly learn features and {\em all}
the attribute label classifiers from class imbalanced training data
in an {\em end-to-end} process. 
Specifically, we introduce {\em incremental minority class discrimination learning}
by formulating a Class Rectification Loss (CRL) regularisation.
The CRL imposes an additional batch-wise class balancing on top of the
cross-entropy loss so to rectify model learning bias due to the
over-representation of the majority classes by promoting
under-represented minority classes 
(Fig. \ref{fig:pipeline}).
%
%
%

\subsection{Limitations of Cross-Entropy Classification Loss}
Convolutional Neural Networks (CNN) are designed to take inputs as
two-dimensional images for recognition tasks \cite{lecun1998gradient}.
For learning a multi-class (per-label) classification CNN model 
(details in ``Network Architecture'',
Sections~\ref{sec:eval_face_attributes}, \ref{sec:eval_clothing_attr},
and \ref{sec:exp_cifar100}), 
the Cross-Entropy (CE) loss function is commonly used 
by firstly predicting the $j$-th attribute posterior probability of $\bm{I}_i$
over the ground truth $a_{i,j}$:
\begin{equation}
{p}(y_{i,j} = a_{i,j} | \bm{x}_{i,j}) = \frac{\exp(\bm{W}_{j}^{\top} \bm{x}_{i,j})} {\sum_{k=1}^{|Z_{j}|} \exp(\bm{W}_{k}^{\top} \bm{x}_{i,j})}
\end{equation}
where $\bm{x}_{i,j}$ refers to the feature vector of $\bm{I}_i$ for the $j$-th attribute label,
and $\bm{W}_k$ is the corresponding classification function parameter. Then compute
the overall loss on a mini-batch of $n_\text{bs}$ images as
the average additive summation of attribute-level loss with equal weight
over all labels:
\begin{equation}
\mathcal{L}_\text{ce} = - \frac{1}{n_\text{bs}}\sum_{i=1}^{n_\text{bs}}  \sum_{j=1}^{n_\text{attr}} \log \Big(p(y_{i,j}=a_{i,j}|\bm{x}_{i,j}) \Big)
\label{eq:loss}
\end{equation}

\begin{figure}[t]
	\centering
	\includegraphics[width=1\linewidth]{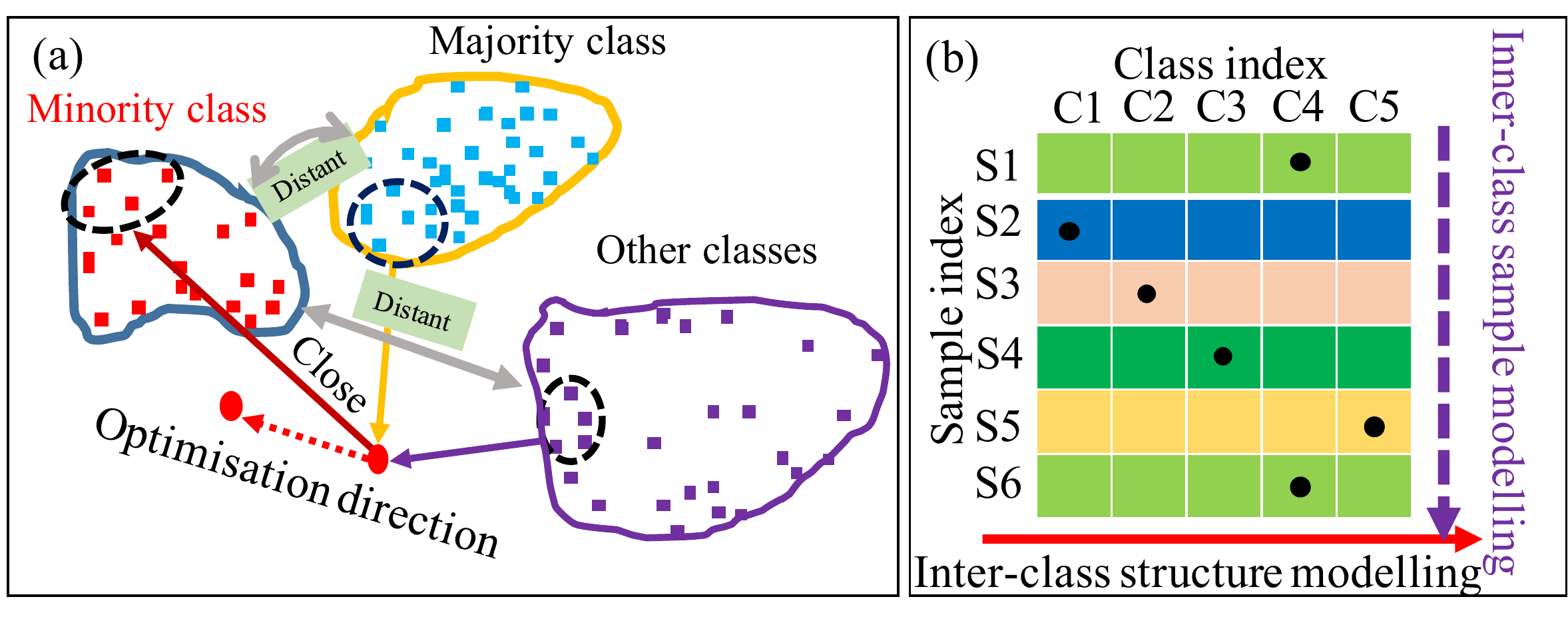}
	\vskip -0.5cm
	\caption{
	Illustration of {\bf (a)}
	inter-class structure rectification
	around decision boundary in the CRL,
	as complementary to
	{\bf (b)} the cross-entropy loss 
	with single-class independent modelling (indicated by dashed arrow).
	}
	\vspace{-0.3cm}
	\label{fig:reason}
\end{figure}

{By design, 
	the cross-entropy loss 
	enforces model learning to respect two conditions:  
	(1) The same-class samples should have class distributions with the identical peak position corresponding to the groundtruth one-hot label.
	(2) Each class corresponds to a different peak position in the class distribution.
	As such, the model is supervised end-to-end to separate the class boundaries
	{\em explicitly} in the prediction space
	and {\em implicitly} in the feature space
	by some in-between linear or nonlinear transformation.
	The CE loss minimises the amount of training error
	by assuming that individual samples and classes are {\em equally} important. 
	To achieve model generalisation with 
	discriminative inter-class boundary separation, 
	it is necessary to have a large training 
	set with sufficiently {\em balanced} class distributions
	(Fig. \ref{fig:feat_visualization}(a)).


	However, given highly class imbalanced training data,
    e.g. X-Domain benchmark,
	model learning by the conventional cross-entropy loss is suboptimal.
	The model suffers from generalising inductive decision boundaries biased towards
	majority classes with ignorance 
	on minority classes (Fig. \ref{fig:feat_visualization}(b)).
	%
	To address this problem, we reformulate the 
	learning objective loss function by {\em explicitly}
	imposing structural discrimination of minority classes against
	others, i.e. {\em inter-class geometry structure modelling} 
	(Fig. \ref{fig:reason}).
	This stresses the structural significance of minority classes in model learning, 
	orthogonal and complementary to 
	the uniform {\em single-class independent modelling} 
	enforced by the cross-entropy loss (Fig. \ref{fig:reason}(b)).
	%
	%
	Conceptually, this design may bring simultaneous benefits to majority class boundary
        learning as shown in our experimental evaluations
        (Tables \ref{tab:arts_face} and \ref{tab:arts_clothing}).
	%
}
\subsection{Minority Class Hard Sample Mining}
\label{sec:method_hard_mining}
{
We explore a hard sample mining strategy 
to enhance minority class manifold rectification by
selectively ``borrowing'' majority class samples from
class decision boundary marginal (border) regions.
Specifically, we estimate minority class neighbourhood structure
by mining {\em both} hard-positive
and hard-negative samples for every selected minority class in {\em each mini-batch} of
training data\footnote{
	We {\em only} consider those minority classes having at least two
	sample images or more in each batch, i.e. ignoring those minority classes having only one sample
	image or none. This enables a more flexible loss function selection, e.g. triplet loss functions
	which typically requires at least two matched samples.
}. 
Our idea is to rectify {\em incrementally} the per-batch class distribution bias
of multi-labels in model learning.
Hence, each improved intermediate model from per-batch training is less inclined towards the 
over-sampled majority classes and more discriminative to the 
under-sampled minority classes (Fig. \ref{fig:feat_visualization}(c)). 
%
Unlike LMLE \cite{huang2016learning} which aims to  
preserve the local structures of {\em both} majority and minority
classes by global clustering of and sampling from the entire training data, 
our model design aims to enhance progressively minority class discrimination by
incremental projective structure refinement.
This idea is inherently compatible with {\em batch-wise} hard-positive
and hard-negative sample mining along the model training trajectory. 
This eliminates the LMLE's drawback in assuming that local group
structures of 
all classes
can be estimated reliably 
by offline {\em global} clustering before model
learning.
}

\vspace{0.1cm}
\noindent {\bf Incremental Batch-Wise Class Profiling }
\label{sec:profile_minority_cls}
For hard sample mining, we first profile the minority and majority classes per label in
each training mini-batch with
$n_\text{bs}$ training samples.
We profile the class distribution $\bm{h}^j = [h_1^j, \dots, h_k^j, \dots h_{|Z_j|}^j]$ 
over $Z_j$ class for each attribute (label) $j$,
where $h_k^j$ denotes the number of training samples 
with the $j$-th attribute value assigned to class $k$.
We then sort $h_k^j$ in the descent order.
As such, we define the minority classes for attribute label $j$ in this {\em mini-batch}
as the smallest classes \sgg{$C_\text{min}^j$}
subject to: 
\begin{equation}
\sum_{k \in C_\text{min}^j} h_k^j \leq \rho \cdot n_\text{bs}
\label{eqn:minority_cls_definition}
\end{equation}
%
%
\sgg{In the most studied
two-class setting \cite{he2009learning}, 
the minority (majority) class is defined as
the one with fewer (more) samples,
i.e. under (above) 50\%. 
However, to our best knowledge there is no standard definition for the
multi-class case.
%
For the definition in Eqn.~\eqref{eqn:minority_cls_definition}
being conceptually consistent to the two-class setting,
we also set $\rho$=50\%.
%
This means that {\em all} minority classes {\em collectively} account for {\em at
	most half or less} samples per batch.}
%
The remaining classes are deemed as the majority classes.
We analysed the effect of choosing different $\rho$ values on model performance (Table \ref{tab:minority_cls_definition}).

\begin{figure}[t]
	\centering
	\includegraphics[width=1\linewidth]{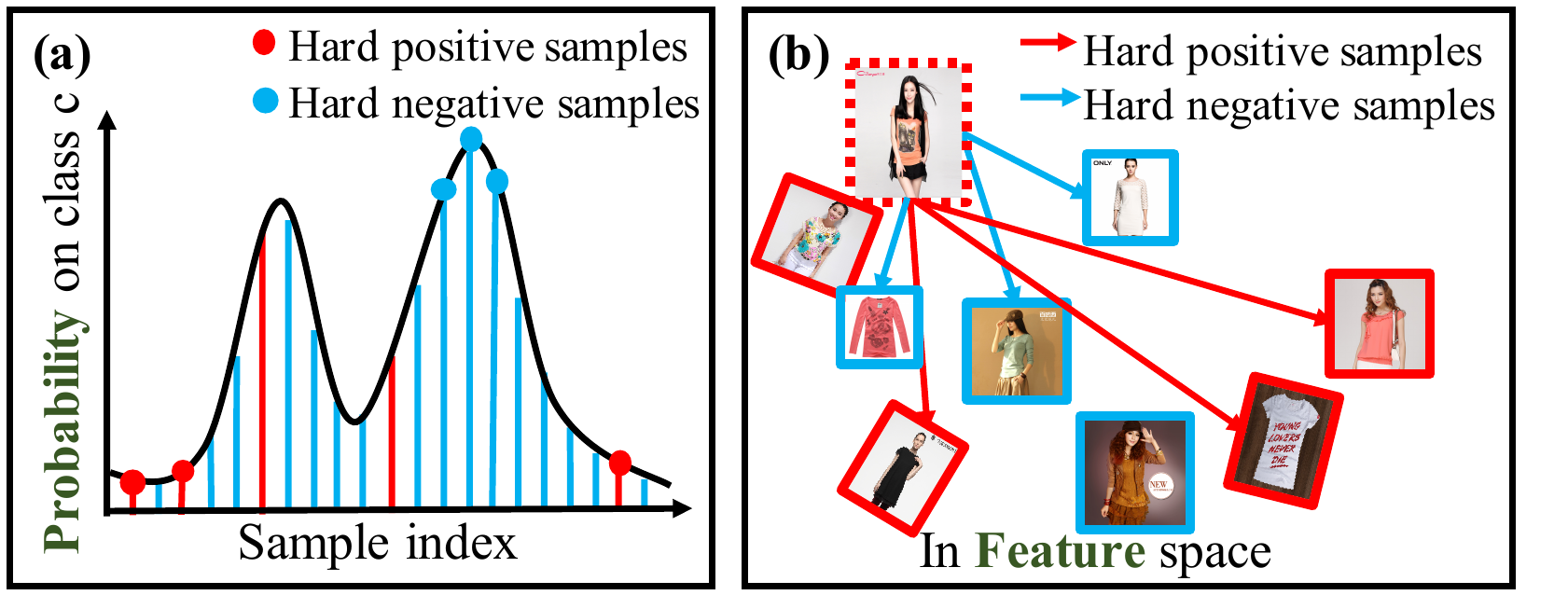}
	\vskip -0.4cm
	\caption{{Illustration of minority class hard sample mining.
	{\bf (a)} {\em Class-level mining}: 
	For each minority class $c$, hard-positives are those samples
	from class $c$ but with low class prediction scores on $c$
	by the current model (red solid circles).
	Hard-negatives are those with high class $c$ prediction scores but on the
        wrong class (blue solid circles).
	{\bf (b)} {\em Instance-level mining}: For each 
        sample
        (dotted red box) of a minority class $c$, hard-positives are
        the samples of class $c$ (solid red box) further away from
        the given sample 
        of class $c$ in the feature space (pointed by a red arrow).
		Hard-negatives are those close to 
		the given sample but from different classes
        (pointed by blue arrow).
	Top-$3$ hard positive and negative samples are shown in 
	the two examples for conceptual illustration.}
	}	
	\vspace{-0.3cm}
	\label{fig:hardmining}
\end{figure}

Given the minority classes, we then consider hard mining therein
at two levels:
class-level (Fig. \ref{fig:hardmining}(a)) and
instance-level (Fig. \ref{fig:hardmining}(b)).
Let us next define the ``hardness'' metrics, hard samples and their
selection.

\vspace{0.1cm}
\noindent {\bf Hardness Metrics }
For hard sample mining, it is necessary to have quantitative metrics
	for ``hardness'' measurement.
Two metrics are considered:
(1) {\em Score based}: 
A model's class prediction score, suitable for class-level hard mining.
(2) {\em Feature based}: The feature distance between data points,
suitable for instance-level hard mining.

 

\vspace{0.1cm}
\noindent {\bf Class-Level Hard Samples } 
At the class-level, we quantify the sample hardness regarding 
a given class per label.
Particularly, for any minority class $c$ of the attribute label $j$, 
we refer ``hard-positives'' to the images $\bm{x}_{i,j}$ of class $c$
($a_{i,j}\! = \! c$ with $a_{i,j}$ the
groundtruth class of the attribute $j$)
given {\em low} prediction scores $p(y_{i,j}\!=\!c|\bm{x}_{i,j})$ on class $c$ 
by the thus-far model, i.e. {\em poor} recognitions. Conversely, by ``hard-negatives'', we refer to the images $\bm{x}_{i,j}$ of 
other classes ($a_{i,j}\! \neq \! c$) 
given {\em high} prediction scores on class $c$ 
by thus-far model, 
i.e. {\em obvious} mistakes. 
Formally, we define them as:
\begin{equation}
\mathcal{P}^\text{cls}_{c,j} = \{ \bm{x}_{i,j} | 
a_{i,j}=c, \; 
\text{low } p(y_{i,j}=c|\bm{x}_{i,j})
 \}
\end{equation} 
\begin{equation}
\mathcal{N}^\text{cls}_{c,j} = \{ \bm{x}_{i,j} | 
a_{i,j} \neq c, \;  
\text{high } p(y_{i,j}=c|\bm{x}_{i,j})  
 \}
\end{equation}
where 
$\mathcal{P}^\text{cls}_{c,j}$ and $\mathcal{N}^\text{cls}_{c,j}$ denote the hard positive and negative sample sets of a minority class $c$ of the attribute label $j$.

\vspace{0.1cm}
\noindent {\bf Instance-Level Hard Samples }
At the instance-level, we quantify the sample hardness regarding
any specific sample instance $\bm{x}_{i,j}$ (groundtruth class $a_{i,j} = c$) 
from each minority class $c$ of the attribute label $j$.
Specifically, we define ``hard-positives'' of $\bm{x}_{i,j}$ as those class $c$ images
$\bm{x}_{k,j}$ (groundtruth class ${a}_{k,j}\! = \! c$) 
by thus-far model with {\em large} distances (low
matching scores) from $\bm{x}_{i,j}$ in the feature space.
In contrast, we define ``hard-negatives'' as those images $\bm{x}_{k,j}$
not from class $c$ 
($a_{k,j} \! \neq \! c$) 
with {\em small} distances (high matching scores) to $\bm{x}_{i,j}$ in
the feature space. We formally define them as: 
\begin{equation}\small
\mathcal{P}_{i,c,j}^\text{ins} = \{ \bm{x}_{k,j} | 
a_{k,j}=c, \; 
\text{large } \mbox{dist}(\bm{x}_{i,j},\bm{x}_{k,j})
\}
\end{equation} 
\begin{equation}\small
\mathcal{N}_{i,c,j}^\text{ins} = \{ \bm{x}_{k,j} |
a_{k,j} \neq c, \;
\text{small } \mbox{dist}(\bm{x}_{i,j},\bm{x}_{k,j})
\}
\end{equation}
where 
$\mathcal{P}_{i,c,j}^\text{ins}$ and $\mathcal{N}_{i,c,j}^\text{ins}$ are the hard positive and
negative sample sets w.r.t. sample instance $\bm{x}_{i,j}$ of minority class $c$ in the 
attribute label $j$, $\mbox{dist}(\cdot)$ is the Euclidean distance metric.

%

\vspace{0.1cm}
\noindent {\bf Hard Mining }
Intuitively, mining hard-positives enables the model to discover and expand sparsely
sampled minority class boundaries, whilst mining hard-negatives aims
to efficiently improve the margin structures of minority class boundary corrupted
by visually similar {\em distracting} classes. 
%
To facilitate and expedite model training, we adopt 
the top-$\kappa$ hard samples mining (selection) strategy.
Specifically, at training time, 
for a minority class $c$ of attribute label $j$
(or a minority class instance $\bm{x}_{i,j}$) in
each training batch data,
we select $\kappa$ hard-positives as the bottom-$\kappa$ scored on $c$ (or
bottom-$\kappa$ (largest) distances to $\bm{x}_{i,j}$),
and $\kappa$ hard-negatives as the top-$\kappa$ scored on $c$ (or top-$\kappa$
  (smallest) distance to $\bm{x}_{i,j}$),
given the current model (or feature space).

\vspace{0.1cm}
\noindent {{\bf Remarks }} 
The proposed hard sample mining strategy 
encourages model learning to concentrate particularly on 
either {\em weak} recognitions or {\em obvious} mistakes
when discriminating sparsely sampled class margins of
the minority classes. 
In doing so, the overwhelming bias towards the majority classes
in model learning is mitigated by {\em explicitly}
stressing minority class discriminative boundary characteristics.
%
To avoid useful information 
of unselected ``easier'' data samples
being lost, we perform scalable hard sample mining 
{\em independently} in each mini-batch during model training and {\em
  incrementally} so over successive mini-batches.
As a result, all training samples are utilised randomly
in the full learning cycle.
%
%
Our model can facilitate naturally both class prediction score
and instance feature distance based matching.
The experiments show that class score rectification
yields superior performance due to 
a better compatibility effect with the score based cross-entropy loss.



\subsection{Minority Class Neighbourhood Rectification}
\label{sec:method_CRL_function}
We introduce a Class Rectification Loss (CRL) regularisation
$\mathcal{L}_\text{crl}$ to rectify model learning bias of the
standard CE loss (Eqn.~\eqref{eq:loss}) due to class imbalanced
training data. This is achieved by incrementally reinforcing the
minority class decision boundary margins with CRL aiming to discover
latent class boundaries whilst maximising their discriminative margins
either directly in the decision score space or indirectly in the
feature space. 
%
%
We design the CRL regularisation
by the learning-to-rank principle
\cite{chopra2005learning,liu2009learning,Ustinova2016hist}
specifically on the minority class hard samples, and re-formulate the
model learning objective loss function Eqn.~\eqref{eq:loss} as:
%
%

\begin{equation}
\mathcal{L}_\text{bln}\ =\ 
\alpha \, \mathcal{L}_\text{crl} + (1-\alpha) \, \mathcal{L}_\text{ce}, 
\quad 
\alpha = \eta \, \Omega_\text{imb}
\label{eq:final_loss}
\end{equation}
where $\alpha$ is a parameter designed to be linearly proportional to
a training {\em class imbalance measure}
$\Omega_\text{imb}$.
Given different individual class data sample sizes,
we define $\Omega_\text{imb}$ as
the minimum percentage count of data samples required over all classes in
order to form an overall uniform (i.e. balanced) class distribution
in the training data. 
%
Eqn.~\eqref{eq:final_loss} imposes an
imbalance-adaptive learning mechanism in CRL regularisation
-- more weighting is assigned to
more imbalanced labels\footnote{Multi-label
  multi-class, e.g. an attribute label has 6$\sim$55 classes.},
whilst less weighting for less imbalanced labels.
Moreover, $\eta$ is 
independent of the per-label imbalance, 
therefore a model hyper-parameter
estimated by cross-validation (independent of individual class imbalance).
In this study, we explore three loss criteria for $\mathcal{L}_\text{crl}$
at {\em both} class-level and instance-level.

\vspace{0.1cm}	
\noindent {\bf (I) Relative Comparison }
First, we consider the seminal triplet ranking loss 
  \cite{liu2009learning} to model the 
  relative relationship constraint between 
intra-class and inter-class.
Considering the small number of training samples in minority classes,
it is sensible to make full use of them in order to effectively handle the 
underlying learning bias.
Hence, we regard each minority class sample 
as an ``anchor'' in the triplet construction to 
compute the batch loss balancing regularisation. 
%

Specifically, for each anchor sample $\bm{x}_{a,j}$, we first construct
a set of triplets based on the mined top-$\kappa$ hard-positives and 
hard-negatives associated with either the corresponding class $c$
of attribute label $j$ (for class-level hard miming), or the sample instance
itself $\bm{x}_{a,j}$ (for instance-level hard mining).
In this way, we form at most $\kappa^2$ triplets $T = \{(\bm{x}_{a,j}, \bm{x}_{+,j}, \bm{x}_{-,j})_s\}_{s=1}^{\kappa^2}$ 
w.r.t. $\bm{x}_{a,j}$, 
and a total of at most $|X_\text{min}| \times \kappa^2$
triplets $T$ for all the anchors $X_\text{min}$ across all the minority classes
of every attribute label. 
We then formulate the following triplet ranking loss to impose 
a CRL class balancing constraint: 
%
%
%
%
\begin{equation} 
\small
\mathcal{L}_\text{crl} = \frac{\sum_{T}
	\max \Big( 0, \,\, m_j +
	\mbox{d}(\bm{x}_{a,j},\bm{x}_{+,j}) - \mbox{d}(\bm{x}_{a,j},
	\bm{x}_{-,j}) \Big) }
{|T|}
%
%
\label{eq:bt}
\end{equation}
{where $m_j$ denotes the class margin of attribute $j$ 
and $\mbox{d}(\cdot)$ is the distance between two samples. 
We consider both class-level and instance-level model
learning rectifications\footnote{\sgg{The maximum operation in
    Eqn.~\eqref{eq:bt} is implemented by a ReLU (rectified linear
    unit) in TensorFlow.}}.

For {\em class-level} rectification, we consider the model predictions
between matched and unmatched pairs:
\begin{equation} \small
\mbox{d}(\bm{x}_{a,j},\bm{x}_{+,j}) = |p_{a,j} - p_{+,j}|, \;\;\;\;
\mbox{d}(\bm{x}_{a,j},\bm{x}_{-,j}) = p_{a,j}-p_{-,j}
\label{eq:cls_pred_diff_pos}
\end{equation} 
where $p_{*,j}$ denotes the model prediction score of $\bm{x}_{*,j}$ 
on the target minority class $c$ of attribute label $j$, with $*\!\in\! \{a,+,-\}$.
The intuition is that, the matched pair is constrained to have similar
prediction scores on the true class (both directions with absolute values), 
higher than that of any negative sample 
by a margin $m_j$ in a single direction (without absolute operation). 
For the triplet ranking, a fixed inter-class margin is often utilised 
\cite{schroff2015facenet} and we set $m_j\!\!=\!\!0.5$
for all attribute labels $j \in \{1,\cdots,n_\text{attr}\}$.
This ensures a correct classification by the maximum a posteriori probability estimation. 

For {\em instance-level} rectification, 
we consider the sample pairwise distance 
in the feature space as:
\begin{equation} \small
\mbox{d}(\bm{x}_{a,j},\bm{x}_{*,j}) = \|\bm{f}_{(a,j)} - \bm{f}_{(*,j)}\|_2,
\label{eq:feat_dist}
\end{equation} 
where $\bm{f}_{(\cdot,j)}$ denotes the attribute $j$ feature vector of the corresponding image sample.
We adopt the Euclidean distance.
In this case, the $m_j$ (Eqn.~\eqref{eq:bt}) specifies the class margin in the feature space. 
%
We apply a geometrically intuitive design: projecting uniformly all the class centres along a unit circle and using the arc length between nearby centres as the class margin. 
That is, we set the class margin for attribute $j$ as:
\begin{equation}
m_j = \frac{2 \pi}{|Z_j| }
\label{eqn:margin}
\end{equation}
where $|Z_j|$ is the number of classes.
}
%

%

\vspace{0.1cm}
\noindent {\bf (II) Absolute Comparison }
Second, we consider the contrastive loss \cite{chopra2005learning} to
enforce absolute pairwise constraints on positive and negative pairs
of minority classes. 
This constraint aims to optimise the boundary of minority classes by
incrementally separating the overlapped (confusing) majority class
samples in batch-wise optimisation.
Specifically, for each sample $\bm{x}_{a,j}$ in a minority class $c$
of an attribute $j$, we use the mined hard samples to build 
positive $P^{+} \!=\! \{\bm{x}_{a,j}, \bm{x}_{+,j}\}$ and negative 
$ P^{-} \! = \! \{\bm{x}_{a,j}, \bm{x}_{-,j}\}$ pairs in each training batch. 
Intuitively, we require the positive pairs to be close 
whilst the negative pairs to be far away in either model score or sample feature space. Thus, we define the CRL as: 
\begin{equation}%
\small
%
%
\begin{split}
\mathcal{L}_\text{crl} = \frac{1}{2} &
\Bigg(\frac{1}{|P^{+}|}\sum_{P^{+}} \mbox{d}(\bm{x}_{a,j}, \bm{x}_{+,j})^2 \;\; + \;\; \\
& \frac{1}{|P^{-}|}\sum_{P^{-}}\max \Big(m_\text{ac} - 
\text{d}(\bm{x}_{a,j}, \bm{x}_{-,j}), \;\; 0 \Big)^2 \Bigg)
\end{split}
\end{equation}
where $m_\text{ac}$ is the between-class margin,
which can be set theoretically to an arbitrary positive number \cite{chopra2005learning}.
We compute the average loss separately for positive and negative
sets to balance their importance even given different sizes.

For {\em class-level} rectification, 
	we consider the model prediction scores of pairs as defined in 
	Eqn.~\eqref{eq:cls_pred_diff_pos}. 
	We set $m_\text{ac}\!\! =\!\!0.5$ to encourage correct prediction.
	For {\em instance-level} rectification, 
	we use the Euclidean distance in the feature space (Eqn.~\eqref{eq:feat_dist})
	for pairwise comparison.
	We empirically set $m_\text{ac}\!\!=\!\!1$, which
	gives satisfactory converging speed and stability in our experiments.

%
%
%
%

\vspace{0.1cm}
\noindent {\bf (III) Distribution Comparison }
Third, we formulate class rectification for minority classes
by modelling the {\em distribution} relationship of positive and negative pairs
(built as in ``Absolute Comparison'').
This distribution based CRL aims to guide model learning by
mining minority class decisive regions {\em non-deterministically}.
In spirit of \cite{Ustinova2016hist}, we represent the distribution of positive ($P^{+}$)
and negative ($P^{-}$) pair sets with histograms $H^{+} \!=\! [h^{+}_1,\cdots,h^{+}_{\tau}]$ and 
\sgg{$H^{-} \!= \![h^{-}_1,\cdots,h^{-}_{\tau}]$} of $\tau$ uniformly spaced bins $[b_1,\cdots,b_{\tau}]$.
We compute the positive histogram $H^{+}$ as:
\begin{equation}
h^{+}_t = \frac{1}{|P^{+}|} \sum_{(i,j) \in P^{+}} \varsigma_{i,j,t}
\end{equation}
where
\begin{equation} \small
	\varsigma_{i,j,t} \! = \! \begin{cases}
	\frac{\text{d}(\bm{x}_{a,j}, \bm{x}_{+,j}) - b_{t-1}}{\Delta}, \; \text{if} \; \mbox{d}(\bm{x}_{a,j}, \bm{x}_{+,j})  \in [b_{t-1},b_{t}] \\
	\frac{b_{t+1} - \text{d}(\bm{x}_{a,j}, \bm{x}_{+,j})}{\Delta}, \; \text{if} \; \mbox{d}(\bm{x}_{a,j}, \bm{x}_{+,j})  \in [b_{t},b_{t+1}] \\ 
	0. \quad\quad\quad\quad\quad\quad\quad\quad\;\;\;  \text{otherwise}
	\end{cases}
\end{equation}
and $\Delta$ defines the pace length between two adjacent bins.
The negative histogram $H^{-}$ can be constructed similarly.
To make minority classes distinguishable from majority classes,
we enforce the two histogram distributions as disjoint as possible.
Formally, we define the CRL regularisation by how much overlapping between
the two distributions:
\begin{equation}
\mathcal{L}_\text{crl} = \sum_{t=1}^{\tau} \big( h^{+}_t \sum_{k=1}^{t} h_k^{-} \big)
\end{equation}
Statistically, this CRL distribution loss estimates the probability that 
the distance of a random negative pair is smaller than that of a random
positive pair, either in the score space or the feature space.
Similarly, we consider 
  both {\em class-level} (Eqn.~\eqref{eq:cls_pred_diff_pos})
  and {\em instance-level} (Eqn.~\eqref{eq:feat_dist}) rectification.

In our experiments (Sec. \ref{exp}),
we compared all six CRL loss designs. By default we deploy the class-level
Relative Comparison CRL in our experiments if not stated otherwise. %

\begin{table*} 
	\footnotesize
	\centering 
	\setlength{\tabcolsep}{0.1cm}
	\caption{
		Statistics of the three datasets utilised in our evaluations. 
	}
	\vskip -0.4cm
	\label{tab:dataset_stat}
	\begin{tabular}{c||c|c|c|c|c|c}
		\hline
		Dataset & Semantics & Labels & Classes & Total Images & Training Images & Test Images \\
		\hline \hline
		CelebA \cite{liu2015deep} 
		& Facial Attribute & Multiple (40) & Binary (2) & 202,599 & 162,770 (3,713$\sim$135,779/class) & 19,867 (432$\sim$17,041/class) \\ \hline
		X-Domain \cite{chen2015deep} 
		& Clothing Attribute & Multiple (9) & Multiple (6$\sim$55) & 245,467 & 165,467 (13$\sim$132,870/class) & 80,000 (4$\sim$64,261/class) \\ \hline
		CIFAR-100 \cite{krizhevsky2009learning} 
		& Object Category & Single (1) & Multiple (100) & 60,000 & 50,000 (500/class) & 10,000 (100/class) \\
		\hline
	\end{tabular}
	\vspace{-0.3cm}
\end{table*}

\vspace{0.1cm} 
\noindent {\bf Further Remarks } 
%
We do not consider exemplars as anchors from majority classes in CRL
because the conventional CE
loss can already model the majority classes well given their frequent sampling.
As demonstrated in our experiments, additional rectification on
majority classes gives some benefit but focusing {\em only} on
minority classes makes the CRL model more cost-effective (Table \ref{tab:cls_mining}).
%
Due to the batch-wise design, the class balancing effect by our proposed regularisor
is incorporated throughout the whole training process progressively.
Conceptually, our CRL shares a similar principle to
Batch Normalisation \cite{ioffe2015batch} in achieving
learning scalability.

\section{Experiments}
\label{exp}


\noindent {\bf Datasets }
For evaluations, we used both imbalanced and balanced benchmark datasets.
Given Table~\ref{tab:dataset_imbalance}, 
we selected the CelebA \cite{liu2015deep}, 
X-Domain \cite{chen2015deep}, and 
CIFAR-100 \cite{krizhevsky2009learning} 
(see Table \ref{tab:dataset_stat} for statistics)
due to:
{\bf (1)} The CelebA provides a {\em class imbalanced learning}
test on multiple {\em binary-class} facial attributes 
with imbalance ratios up to 1:43. 
Specifically, it has 202,599 web in-the-wild images
from 10,177 person identities with on average 20 images per person. 
Each image is annotated by 40 attribute labels.
Following \cite{liu2015deep,huang2016learning}, we used 162,770 images for model training (including 10,000 images for validation),
and the remaining 19,867 for test.
{\bf (2)} The X-Domain
 offers an {\em extremely class imbalanced learning} test on multiple
{\em multi-class} clothing attributes with 
the imbalance ratios upto 1:4,162. 
This dataset consists of 245,467 shop images 
extracted from online retailers.
Each image is annotated by 9 attribute labels.
Each attribute has a different set of mutually exclusive class values,
sized from 6 (``sleeve-length'') to 55 (``colour'').
In total, there are 178 distinctive attribute classes over the 9 labels.
We randomly selected 165,467 images for training (including 10,000 images for validation) and the remaining 80,000 for test.
{\bf (3)} The CIFAR-100 provides a {\em single-label class
balanced learning} test. 
This benchmark contains 100 classes with each having 
600 images. 
This test provides a complementary evaluation of the proposed method
against a variety of benchmarking methods, 
and moreover, facilitates extra in-depth model analysis 
under simulated class imbalanced settings. We used the standard 490/10/100 training/validation/test split
per class \cite{krizhevsky2009learning}.
\noindent {\bf Performance Metrics }
The classification accuracy \cite{dong2016multi,liu2015deep} 
that treats all classes uniformly 
is not appropriate 
for class imbalanced test, as
a naive classifier that predicts every test sample 
as majority classes can still achieve a high overall accuracy although
it fails all minority class samples.
Since we consider the multi-class imbalanced classification test,
the common true/false (positive/negative) rates for binary-class 
classification are no longer valid.
In this work, we adopt the {\em sensitivity} measure
that leads to a {\em class-balanced} accuracy 
by considering particularly the class distribution statistics
\cite{fernandez2011dynamic} and
%
generalises the conventional binary-class criterion
\cite{huang2016learning}.
Formally, we compute the per-class sensitivity based on the classification
confusion matrix 
as:
\begin{equation}
S_i = \frac{n_{(i,{i})}}{n_i}, \;\; 
n_i = \sum_{j=1}^c n_{(i, {j})}, \;\; i \in \{1,2,\cdots,c\}
\label{eq:sensivity}
\end{equation}
where $n_{(i,{j})}$ is the number of class $i$ test samples predicted
by a model as class $j$,
and $n_i$ is the size of class $i$ (totally $c$ classes). 
Therefore, the confusion matrix diagonal refers to
correctly classified sample numbers of individual classes whilst 
the off-diagonal to the incorrect numbers.
We define the {\em class-balanced accuracy} (i.e. mean sensitivity) as: 
\begin{equation}
A_\text{bln} = \frac{1}{c}\sum_{i=1}^c S_i
	\label{eq:accuracy}
\end{equation}
%
The above metric is for the single-label case.
For the multi-label test,
we average the mean sensitivity measures over all labels (attributes)
to give the overall class-balanced accuracy.
\begin{table*} 
	\scriptsize
	\centering
	\renewcommand{\arraystretch}{0.85}
	\setlength{\tabcolsep}{0.21cm}
	\caption{
		Facial attribute recognition on the CelebA benchmark \cite{liu2015deep}. 
			``*'': Class imbalanced learning models. 
			{Metric}: {\em Class-balanced accuracy} (\%).
			Instance level hard mining. The $1^\text{st}$/$2^\text{nd}$
			best results are indicated in red/blue.
			MthOpen: Mouth Open;
			HighChb: High Cheekbones;
			HvMkup: Heavy Makeup;
			WvHair: Wavy Hair;
			OvFace: Oval Face;
			PntNose: Pointy Nose;
			ArEyeb: Arched Eyebrows;
			BlkHair: Black Hair;
			StrHair: Straight Hair;
			BrwHair: Brown Hair;
			BldHair: Blond Hair;
			GrHair: Gray Hair;
			NrwEye: Narrow Eyes;
			RcdHl: Receding Hairline;
			5Shdw: 5 o’clock Shadow;
			BshEb: Bushy Eyebrows;
			RsChk: Rosy Cheeks;
			DbChn: Double Chin;
			EyeGls: Eyeglasses;
			SdBurn: Sideburns;
			Mstch: Mustache;
			PlSkin: Pale Skin.
	}
	\vskip -0.4cm
	\label{tab:arts_face}
	\begin{tabular}{c||c|c|c|c|c|c|c|c|c|c|c|c|c|c|c|c|c|c|c|c||c}
		\cline{1-22}
		\backslashbox{\bf Methods}{{\bf Attributes}}&  \rotatebox{90}{Attractive} & \rotatebox{90}{MthOpen}  & \rotatebox{90}{Smiling}  & \rotatebox{90}{Lipstick}  &   \rotatebox{90}{HighChb} & \rotatebox{90}{Male}  &  \rotatebox{90}{HvMkup} &  \rotatebox{90}{WvHair} & \rotatebox{90}{OvFace}  &
		\rotatebox{90}{PntNose}  & \rotatebox{90}{ArEyeb}  & \rotatebox{90}{BlkHair}  &   \rotatebox{90}{Big Lips} & \rotatebox{90}{Big Nose}  &  \rotatebox{90}{Young} &  \rotatebox{90}{StrHair} & \rotatebox{90}{BrwHair} & \rotatebox{90}{EyeBag} & \rotatebox{90}{Earrings} & \rotatebox{90}{NoBeard} & \rotatebox{90}{}   \\ 
		\cline{1-21}
		\bf Imbalance ratio (1:x)&1&   1&   1&   1& 1&   1&   2&   2& 3&   3&   3&   3&3&   3&   4&   4& 4&   4 &   4&   5 \\  
		\hline
		
		Triplet-$k$NN \cite{schroff2015facenet} 
		& 83 & 92 & 92 & 91& 86 & 91 & 88 & 77 & 61 & 61 & 73 & 82 & 55 & 68 & 75 & 63 & 76 & 63 & 69 & 82 &  \\
		
		PANDA \cite{zhang2014panda} 
		& 85 & 93 & \3{98} & \3{97} & \3{89} & \1{99} & {95} & 78 & 66 & 67 & 77 & 84 & 56 & 72 & 78 & 66 & \3{85} & 67 & 77 & 87 &  \\
		
		ANet \cite{liu2015deep}
		& \3{87} & \1{96} & 97 & 95 & \3{89} & \1{99} & \3{96} & \3{81} & {67} & 69 & 76 & \3{90} & 57 & \3{78} & \3{84} & 69 &  83 & 70 & \3{83} & \3{93} &  \\

		DeepID2 \cite{sun2014deep}
		& 78 & 89 & 89 & 92  & 84  & 94  & 88 & 73 & 63 & 66 
		& 77 & 83 & {62} & 73 & 76 & 65 & 79  & 74 & 75  & 88 &  \\ 
		\cline{1-21}
		
		Over-Sampling* \cite{drummond2003c4}
		&77 &89 &90 &92 &84 &95 &87 &70 &63 &67 
		&\3{79} &84 &61 &73 &75 &66 &82 &73 & 76&88 &  \\ 
		
		Down-Sampling* \cite{drummond2003c4}
		&78 &87 &90 &91&80&90&89  &70  &58  &63
		&70 &80 &61 &76&80&61&76  &71  &70  &88&\\ 
		
		Cost-Sensitive* \cite{he2009learning}
		&78 &89 &90 &91&85&93&89  &75  &64 &65
		&78&85&61&74&75&67&84&74&76&88& \\
		
		Threshold-Adjustment* \cite{chen2006decision}
		&69&89&88&89&83&95&89&77&\1{72}&\1{72}
		&76&86&\3{66}&76&24&\1{73}&81&\3{76}&76&15   \\ 
		\cline{1-21} 
		
		LMLE* \cite{huang2016learning} 
		& \1{88} & \1{96} & \1{99} &  \1{99} & \1{92} &  \1{99} &  \1{98} & \1{83} &  \3{68} & \1{72} &\3{79} & \1{92} & 60 & \1{80} & \1{87} & \1{73} & \1{87} & {73} & \3{83} & \1{96} &\\ 
		\cline{1-21}
		
		
		
		{\bf CRL* }  &81  &  \3{94}   & 92  &  95  &  87 &   \3{98}  &  90    &79   & 66&\3{71}&\1{80}    &88&    \1{67}&    77& 83& 72 &84 &\1{79}&\1{84}    &\3{93}&\\
		
		\cline{1-21}
		\hline \hline
		\backslashbox{\bf Methods}{{\bf Attributes}} & \rotatebox{90}{Bangs}& \rotatebox{90}{BldHair} & \rotatebox{90}{BshEb} & \rotatebox{90}{Necklace} &  \rotatebox{90}{NrwEye} & \rotatebox{90}{5Shdw} &  \rotatebox{90}{RcdHl} &  \rotatebox{90}{Necktie} &  \rotatebox{90}{EyeGls} &  \rotatebox{90}{RsChk} & \rotatebox{90}{Goatee} &  \rotatebox{90}{Chubby} &  \rotatebox{90}{SdBurn} &  \rotatebox{90}{Blurry} &  \rotatebox{90}{Hat} & 
		\rotatebox{90}{DbChn} &  \rotatebox{90}{PlSkin} &
		\rotatebox{90}{GrHair}&\rotatebox{90}{Mstch}&\rotatebox{90}{Bald}& \em
		\rotatebox{90}{\bf Mean} \\ 
		\cline{1-21}
		\bf Imbalance ratio (1:x)&6&   6&6&   7&8&   8&  11&  13&14&  14&  15&  16&17&  18&  19&  20&22&  23&  24&  43\\\hline
		
		Triplet-$k$NN \cite{schroff2015facenet} 
		&81& 81 & 68 & 50 & 47 & 66 & 60 & 73 & 82 & 64 & 73 & 64 & 71 & 43 & 84 & 60 & 63 & 72 & 57 & 75 &  \bf{72} \\
		
		PANDA \cite{zhang2014panda} 
		&92& 91 & 74 & 51 & 51 & 76 & 67 & 85 & 88 & 68 & 84 & 65 & 81 & 50 & 90 & 64 & 69 & 79 & 63 & 74 &   \bf{77} \\
		
		ANet \cite{liu2015deep}
		&90& 90 & \3{82} & 59 & 57 & 81 & 70 & 79 & 95 & 76 & 86 & 70 & 79 & 56 & 90 & 68 & 77 & 85 & 61 & 73 &   \bf{80} \\
		
		DeepID2 \cite{sun2014deep}
		&91& 90 & 78 & 70 & 64  & \3{85} & 81  & 83 & 92  & 86 & \3{90}  
		& {81}  & 89 & 74  & 90 & 83 & 81 & 90 & 88 & 93  &  \bf{81}  \\ \hline 
		Over-Sampling* \cite{drummond2003c4}&90&90 &80 &\3{71} &65 &\3{85} &82 &79  &91  &\3{90} &89&\3{83} &\3{90} &76&89&\3{84}&82&90&\3{90}&92& \bf{82}  \\ 
		
		Down-Sampling* \cite{drummond2003c4}
		&88&85 &75 &66 &61&82&79&80  &85  & 82 &85
		&78 &80 &68&90&80&78&88&60&79& \bf{78}   \\  
		
		Cost-Sensitive* \cite{he2009learning}
		&90&89 &79 &\3{71} &65&84&81&82  &91  &\1{92}&86 
		&82&\3{90}&76&90&\3{84}&80&90&88&\3{93}& \bf{82} \\\
		
		Threshold-Adjustment* \cite{chen2006decision}
		&93&92&\1{84}&62&\3{71}&82&\3{83}&76&95&82&89&81&89&\3{78}
		&\3{95}&83&\3{85}&\3{91}&86&\3{93}&{\bf79}   \\ \hline
		
		LMLE* \cite{huang2016learning} 
		& \1{98}& \1{99} &\3{82} & 59 &  59 & 82 & 76 & \1{90} &\3{98} &  78 &\1{95} & 79 & 88 & 59 & \1{99} & 74 & 80 & \3{91} & 73 &  90 &   \3{84} \\ \hline

%
		
		{\bf CRL* (Ours)}
		&\3{95} & \3{95}  &\1{84}  &  \1{73} &   \1{73} &   \1{89}  &  \1{88} &   \3{87} &  \1{99}
		&\3{90}    &\1{95}  &  \1{87}  &  \1{95}  &  \1{86}  &  \1{99}   & \1{89}  &  \1{92} &   \1{96}
		&\1{93}   & \1{99}&\1{\bf87} \\
		\hline
		
	\end{tabular}
	\vspace{-0.2cm}
\end{table*}

\vspace{0.1cm}
\noindent {\bf Imbalanced Learning Methods for Comparison } 
We considered five existing class imbalanced learning methods: 
{\bf (1)} Over-Sampling \cite{drummond2003c4}: 
A multi-label re-sampling strategy to build a
  more balanced set before model learning through over-sampling
  minority classes by random replication. 
%
{\bf (2)} Down-Sampling \cite{drummond2003c4}:
Another training data re-sampling method 
based on under-sampling majority classes with random sample
        removal. 
{\bf (3)} Cost-Sensitive \cite{he2009learning}: 
A class-weighting strategy by assigning greater misclassification penalties to
minority classes and smaller penalties to majority classes in loss design.
We assign the class weight as 
$w_i \! = \! \exp(-r_i)$
where $r_i$ specifies the ratio of class $i$ in training data.
%
{\bf (4)} Threshold-Adjustment \cite{chen2006decision}:
Adjusting the model decision threshold in test time by
incorporating the class probability prior $r_i$, 
\sgg{e.g. moderating the original model prediction ${p}_i$ to 
$\tilde{p}_i \! = \! {p}_i*\exp(-r_i)^T$ where
$T \!\! \in \!\! \{1,2,3,4,5\}$ is a temperature (softening) 
parameter estimated by cross validation.} 
Given 
$\tilde{p}_i$, 
we then use the maximum a posteriori probability for class prediction.
{\bf (5)} 
LMLE \cite{huang2016learning}: 
A state-of-the-art class imbalanced deep learning model
exploiting the class structure for improving minority class
modelling. 
For fair comparisons, 
all the methods were implemented on the same network
architecture (details below),
\sgg{with the parameters set by following
	the authors' suggestions if available or cross-validation.}
All models were trained on the same training data,
and evaluated on
the same test data.
\sgg{We adopted the class-level
	relative comparison CRL for all remaining experiments if not stated otherwise.}

\subsection{Comparisons on Facial Attributes Recognition}
\label{sec:eval_face_attributes} 

\noindent {\bf Competitors } 
{
We compared the proposed CRL model with 9 existing methods including the 5 
class imbalanced learning models above
and other 4 state-of-the-art deep learning models 
for facial attribute recognition on the CelebA benchmark: 
{ (1)} PANDA \cite{zhang2014panda},  
%
{ (2)} ANet \cite{liu2015deep},
%
{ (3)} Triplet-$k$NN \cite{schroff2015facenet},
and { (4)} DeepID2 \cite{sun2014deep}.
}
%
%

\vspace{0.1cm}
\noindent {\bf Network Architecture }
We adopted the 5-layers CNN architecture DeepID2 \cite{sun2014deep} as
the base network for training all class imbalanced learning methods including
CRL and LMLE. 
\sgg{Training DeepID2 was based on the conventional CE loss
  (Eqn.~\eqref{eq:loss}). This provides a baseline for
  evaluations with and without CRL.}
Moreover, the CRL allows multi-task learning in the
  spirit of \cite{evgeniou2004regularized,ando2005framework}, with an
  additional 64-dim FC$_2$ feature layer and 
  a 2-dim binary prediction layer for each face attribute.

\vspace{0.1cm}
\noindent {\bf Parameter Settings } 
We trained the CRL from scratch 
by 
the learning rate at 0.001,
the decay at 0.0005,
the momentum at 0.9,
the batch size at 256, and
the epoch at 921. 
We set the loss weight $\eta$ (Eqn. \eqref{eq:final_loss}) to 0.01.

\vspace{0.1cm}
\noindent {\bf Overall Evaluation } 
Facial attribute recognition performance comparisons are shown
in Table \ref{tab:arts_face}.
It is evident that CRL outperforms
all competitors including the 
attribute recognition models and class imbalanced learning methods
on the overall mean accuracy.
Compared to the best non-imbalanced learning model DeepID2, CRL
improves the average accuracy by $6\%$. 
Compared to the state-of-the-art imbalanced
learning model LMLE, CRL is better {on average accuracy by} $3\%$. 
Other classic
imbalanced learning methods perform inferiorly than both CRL and LMLE. 
In particular,
Over-Sampling brings only marginal gain
with no clear difference across all imbalance degrees,
suggesting that replication 
based data rebalance is
limited in introducing useful information. 
Cost-Sensitive is largely similar to Over-Sampling.
%
The performance drop by Down-Sampling and Threshold-Adjustment is due to 
discarding
useful data in balancing the class data distribution
and imposing potentially inconsistent adjustment
to model prediction. 
%
This shows that (1) not all class imbalanced learning methods are helpful,
and (2) the clear superiority of our batch-wise incremental minority class rectification method
in handling biased model learning over alternative methods. 


\begin{figure} 
\centering
\includegraphics[width=1\linewidth]{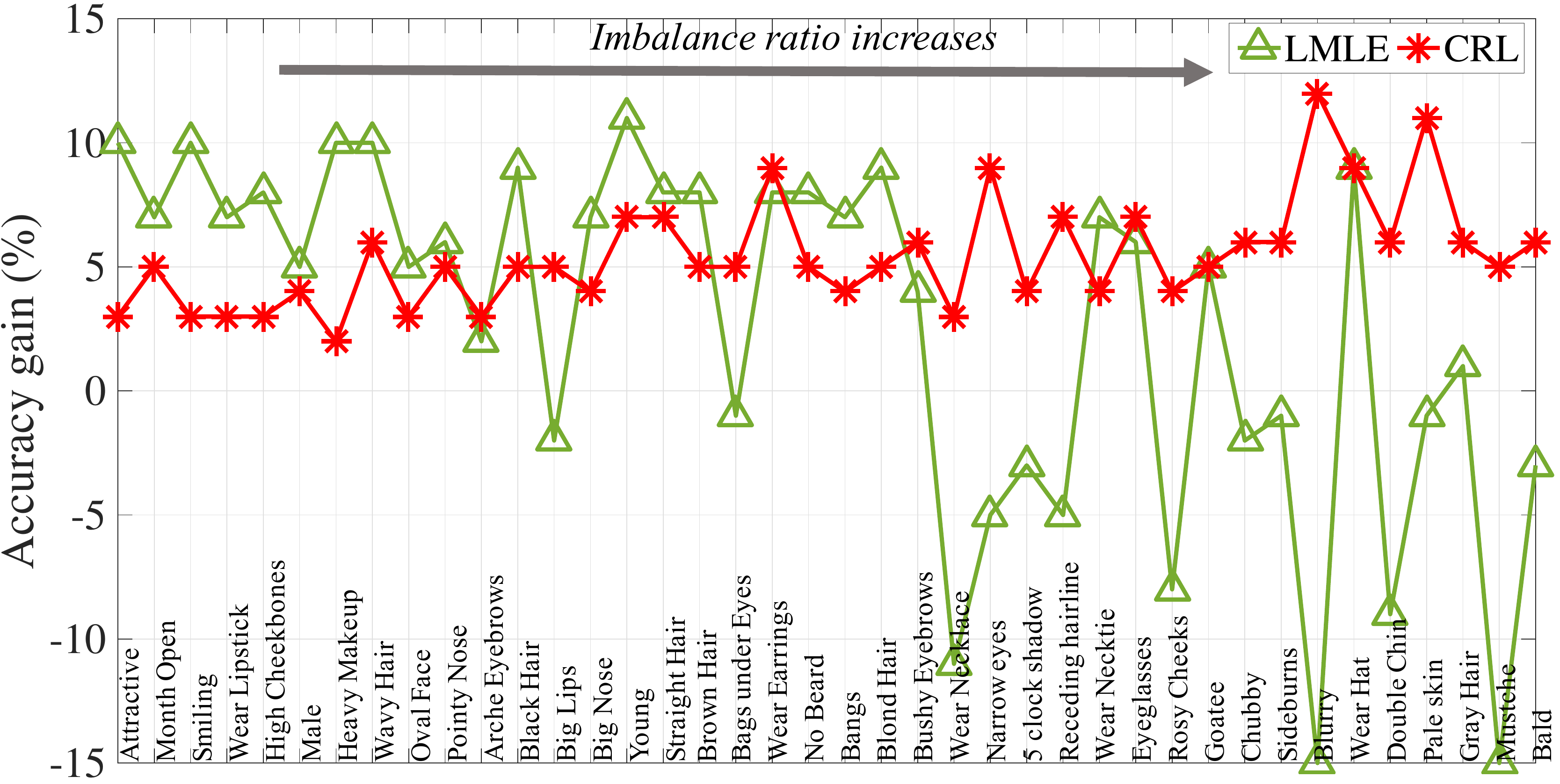}
\vskip -0.4cm
\caption{
	Performance {\em additional} gain over the DeepID2 by
	the LMLE and CRL models
	on
	the 40 CelebA binary-class facial attributes \cite{liu2015deep}.
	Attributes are sorted from left to right
	in increasing class imbalance ratio.
}
\label{fig:accgain_face}
\vspace{-0.4cm}
\end{figure}

\begin{table}[th]
	\footnotesize
	\centering
	\caption{\sgg{Performance {\em additional} gain over the DeepID2 by LMLE and CRL
		on bottom-20 and top-20 CelebA facial attributes in mean {\em class-balanced accuracy} (\%).
		Negative number means performance drop.}
	}
	\vskip -0.3cm
	\label{tab:LMLEvsCRL}
	\begin{tabular}{c||c|c}
		\hline
		Imbalance Ratio
		&  Bottom-20 (1:1$\sim$1:5) & Top-20 (1:6$\sim$1:43)  \\ 
		\hline \hline
		LMLE \cite{huang2016learning} &  \bf +7 & -2  \\ \hline
		\bf CRL&  +5 &\bf +6  \\
		\hline
	\end{tabular}
	\vspace{-0.3cm}
\end{table}

\begin{figure} 
	\centering
	\includegraphics[width=1\linewidth]{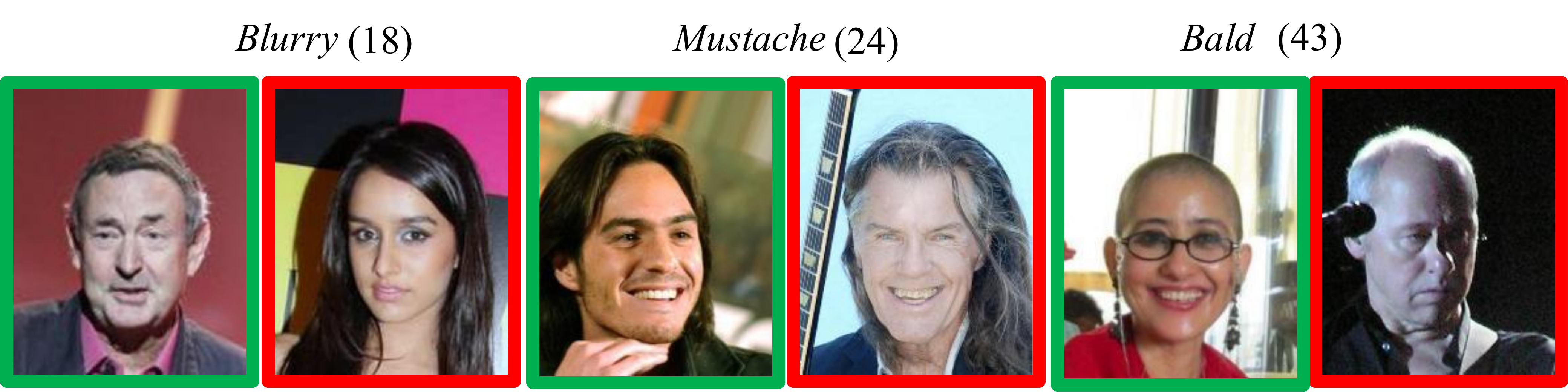}
	\vskip -0.4cm
	\caption{
 	Examples (3 pairs) of facial attribute
		recognition (imbalance ratio in bracket). 
		For each pair (attribute), DeepID2 missed both, whilst CRL
		identified the image with green box but failed the
		image with red box. 
	}
	\label{fig:vis_face}
	\vspace{-0.5cm}
\end{figure}

\vspace{0.1cm}
\noindent {\bf Further Analysis }
We examined the characteristics of model performance on individual
attributes exhibiting different class imbalance ratios.
In particular, we further analysed CRL and the best competitor LMLE 
against the base model DeepID2 without class imbalanced learning.
To that end,
we split the 40 facial attributes into two groups at a 1:5 imbalance ratio:
bottom-20 (the first 20 in Table \ref{tab:arts_face}) and top-20 (the remaining) imbalanced attributes.
Figure~\ref{fig:accgain_face} and Table \ref{tab:LMLEvsCRL} show that:
{\bf (1)} CRL improves the prediction accuracy on all attributes (above ``0''),
whilst LMLE can give weaker prediction than DeepID2 especially on highly
imbalanced attributes. 
This suggests that CRL is more robust in coping with different
imbalanced attributes, especially more extremely imbalanced classes.
{\bf (2)} LMLE is better at the bottom-20 imbalanced attributes,
improving the mean accuracy by 7\% {\em versus} 5\% by CRL.
For instance, 
CRL is outperformed by LMLE 
on the ``Attractive'' (balanced) and ``Heavy Makeup'' attributes by
7\% and 8\%, respectively. 
This suggests that LMLE is better for less-extremely imbalanced
attributes. 
%
{\bf (3)} LMLE performance degrades on top-20 imbalanced attributes by $2\%$ in mean accuracy.
Specifically, LMLE performs worse than DeepID2
on 
most attributes with imbalance ratio greater than 1:7, 
starting from ``Wear Necklace'' in Table~\ref{tab:arts_face}.	
This is in contrast to CRL which 
achieves an even better performance gain at $6\%$ on top-20.
On some very imbalanced attributes, 
CRL outperforms LMLE significantly, e.g. by $20\%$ on
``Mustache" and $27\%$ on ``Blurry". 
Interestingly, the ``Blurry'' attribute is visually
challenging due to its global characteristics not defined by
local features therefore very subtle, similar to the ``Mustache'' attribute
(see Fig.~\ref{fig:vis_face}).
This demonstrates that CRL is
superior and more scalable than LMLE 
in coping with severely imbalanced data learning.
This is due to (1) incremental batch-wise minority class predictive
boundary rectification 
which is independent to global training data class structure, 
and (2) end-to-end deep learning for joint feature and classifier optimisation
which LMLE lacks.

\vspace{0.1cm}
\noindent {\bf Model Training Cost Analysis }
We analysed the model training cost of CRL and LMLE
on a workstation with $1$ NVIDIA Tesla K$40$ GPU and $20$ E5-2680 @ $2.70$GHz CPUs.
For LMLE, we used the codes released by the authors\footnote{
The k-means clustering function 
is not included in the original codes. 
We used the VLFeat's implementation \cite{vedaldi08vlfeat}
with the default setting
as $1,000$ maximum iterations and $10$ repetitions.}
with the original settings
($4$ rounds of training
each with $5,000$ iterations of CNN optimisation).
The training was initialised by pre-trained DeepID2 face recognition features.
%
On our workstation, LMLE took a total of 264.8 hours to train\footnote{
We did not consider the time cost for
pre-training the DeepID2 
(needed for extracting the initial features
for the first round of data pre-processing) on face identity labels from CelebFaces+ \cite{sun2014deep}
due to lacking the corresponding codes and details.
We used the pre-trained DeepID2 model 
thanks to the helpful sharing by the LMLE authors.},
with each round taking 66.2 hours including
24.5 hours for ``clustering+quintuplet construction''
and 
41.7 hours for ``CNN model optimisation''.
In contrast, CRL took 27.2 hours, that is 9.7 (264.8/27.2) times faster than LMLE.

%

%
We further examined model convergence rate.
Specifically, LMLE converges quicker than CRL
on training batch iterations, 
LMLE's 20,000 
{\em versus} CRL's 540,000.
This is reasonable as LMLE benefits uniquely from both a specifically designed
data structural pre-processing (building quintuplets) of the entire training
data which is a computationally expensive procedure, 
and a model pre-training process on auxiliary face recognition labels.
However, LMLE is significantly slower than CRL 
in the overall CNN training time: LMLE's 166.6 hours {\em versus}
CRL's 27.2 hours. 

\begin{table*} [h] 
	\footnotesize
	\centering
	\renewcommand{\arraystretch}{0.85}
	\setlength{\tabcolsep}{0.23cm}
	\caption{Clothing attributes recognition on the X-Domain dataset \cite{chen2015deep}. 
		``*'': Imbalanced data learning models. 
		{Metric}: {\em Class-balanced
		accuracy} (\%).
		Slv-Shp: Sleeve-Shape; 
		Slv-Len: Sleeve-Length.
		The $1^\text{st}/2^\text{nd}$ best results are highlighted in red/blue.
	}
	\vskip -0.4cm
	\label{tab:arts_clothing}
	\begin{tabular}{c||c|c|c|c|c|c|c|c|c||c}
		\hline
		\backslashbox{\bf Methods}{\bf Attributes}
		
		& Category & Colour & Collar & Button  & Pattern & Shape & Length  & Slv-Shp& Slv-Len & \multirow{2}{*}{\bf Mean}\\ \cline{1-10}
		\bf Imbalance ratio (1:x)
		&2&138&210&242&476&2,138&3,401&4,115&4,162 & \\\hline \hline
		DDAN \cite{chen2015deep} 
		&46.12 &31.28&22.44&40.21    &29.54 &23.21&32.22  &19.53&40.21   & {\bf 31.64}\\
		FashionNet \cite{liu2016deepfashion} 
		&48.45 &36.82&25.27&43.85     &31.60 &27.37&38.56&20.53 &45.16   & {\bf 35.29}\\
		
		DARN \cite{huang2015cross} 
		&65.63  &44.20&31.79&58.30 &44.98  & 28.57 &45.10 & 18.88&51.74    &{\bf 43.24}\\
		
		MTCT \cite{dong2016multi} 
		&72.51 &74.68&70.54&76.28&76.34&68.84&77.89& 67.45&77.21  &{\bf 73.53}\\ \hline

		Over-Sampling* \cite{drummond2003c4}
		&73.34 &75.12&\3{71.66}&77.35&77.52&68.98&78.66& 67.90&78.19& {\bf 74.30}  \\  
		
		Down-Sampling* \cite{drummond2003c4}
		&49.21&33.19&19.67&33.11&22.22&30.33&23.27&12.49&13.10&{\bf 26.29}\\
		
		Cost-Sensitive* \cite{he2009learning} 
		&\3{76.07} &\3{77.71}&71.24&\3{79.19}&77.37&69.08&78.08& 67.53&77.17&{\bf 74.49} \\ 
		
		Threshold-Adjustment* \cite{chen2006decision}
		& 72.51 &  75.14  &  71.34  & 77.91  & 77.46  & 70.25  & 78.78  & \3{70.78}   &78.37& {\bf74.72} \\
		\hline
		LMLE* \cite{huang2016learning}&75.90&77.62&70.84&78.67&\3{77.83}&\3{71.27}&\3{79.14}&{69.83}&\3{80.83}&\3{\bf 75.77} \\ \hline
		
		
		
		\bf CRL* (Ours)&\1{77.69}&\1{82.01}	&\1{77.01}	&\1{82.37}	&\1{81.39}	&\1{74.96} &\1{84.81}	&\1{80.48}	&\1{83.02}&\1{\bf80.42} \\
		\hline
	\end{tabular}
	\vspace{-0.2cm}
\end{table*}

\subsection {Comparisons on Clothing Attributes Recognition}
\label{sec:eval_clothing_attr}

\noindent {\bf Competitors }
{Except the five imbalanced learning methods,
we also compared CRL against four other state-of-the-art clothing attribute recognition
models: }
{(1)} 
DDAN \cite{chen2015deep},
{(2)} 
DARN \cite{huang2015cross},
%
{(3)} FashionNet\footnote{We implemented this
	FashionNet without the landmark detection branch 
	since no landmark labels are
	available in the X-Domain dataset.} \cite{liu2016deepfashion},
%
and {(4)} 
MTCT \cite{dong2016multi}.
	%


\vspace{0.1cm}
\noindent {\bf Network Architecture }
We used the same network structure as the MTCT \cite{dong2016multi}. 
Specifically, this network is composited of 
five stacked NIN conv units \cite{lin2013network} and 
$n_\text{attr}$ parallel branches with each 
a 3-FC-layers sub-network for modelling a distinct
attribute respectively,
in the multi-task learning spirit \cite{ando2005framework}.
\sgg{We trained MTCT 
	using the CE loss (Eqn.~\eqref{eq:loss}).}

\vspace{0.1cm}
\noindent {\bf Parameter Settings }
We pre-trained the base network
on ImageNet-1K \cite{russakovsky2015imagenet} at
the learning rate 0.01,
then fine-tuned the CRL model on the X-Domain
images at a lower learning rate 0.001.
We set the decay to 0.0005,
the momentum to 0.9,
the batch size to 128, and
the epoch to 256.
We set the loss weight $\eta$ (Eqn. \eqref{eq:final_loss}) to 0.01.

\vspace{0.1cm}
\noindent {\bf Overall Evaluation }
Table \ref{tab:arts_clothing} shows the comparative evaluation of 10 different
models on the X-Domain benchmark.
It is evident that CRL surpasses all
prior state-of-the-art models on all attribute labels.
This shows the superiority and scalability of 
our incremental minority class
rectification 
in tackling extremely imbalanced attribute data, with the maximum imbalance
ratio 4,162 {\em versus} 43 in CelebA attributes.
For example, CRL surpasses the 
best competitor LMLE by $4.65\%$ in mean accuracy.
Traditional class imbalanced learning methods
behave similarly as on facial attributes,
except that Threshold-Adjustment also
yields a small gain similar as Cost-Sensitive.
Other models without an explicit imbalanced learning mechanism like
DDAN, FashionNet, DARN and MTCT
suffer notably. 
%
%

\begin{table}[th]
	\footnotesize
	\centering
	\caption{\sgg{Performance {\em additional} gain over the MTCT by LMLE and CRL
		on bottom-1 and top-8 X-Domain clothing attributes in mean accuracy (\%).}
	}
	\vskip -0.3cm
	\label{tab:LMLEvsCRL_clothing}
	\begin{tabular}{c||c|c}
		\hline
		Imbalance Ratio
		&  Bottom-1 (1:2) & Top-8 (1:138$\sim$1:4,162)  \\ 
		\hline \hline
		LMLE \cite{huang2016learning} &  +3.39 & +2.10  \\ \hline
		\bf CRL& \bf +5.18 &\bf +7.10  \\
		\hline
	\end{tabular}
	\vspace{-0.2cm}
\end{table}

\begin{figure} 
	\includegraphics[width=1\linewidth]{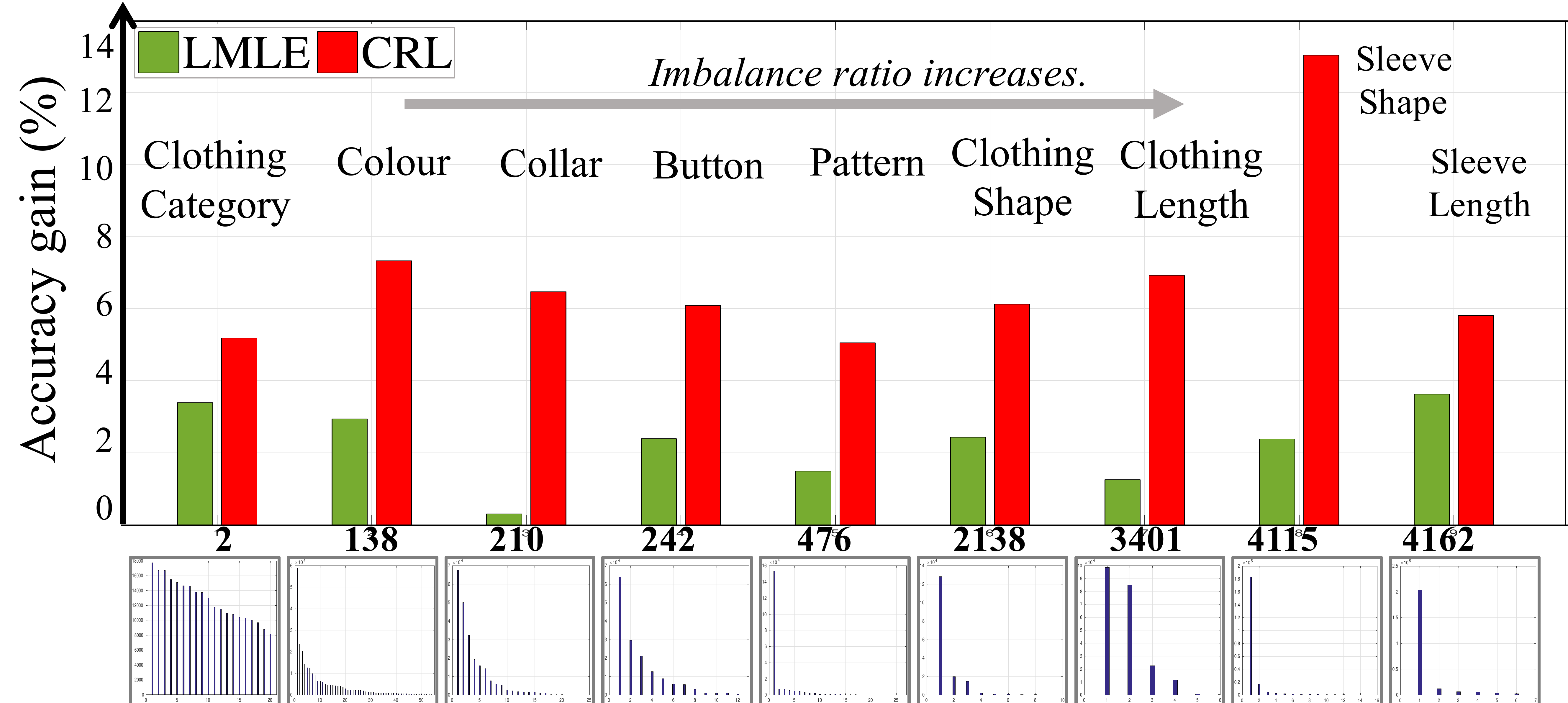}
	\vskip -0.3cm
	\caption{
		Performance {\em additional} gain over the MTCT 
		by the LMLE and CRL models on 9 X-Domain multi-class clothing attributes
		\cite{chen2015deep} with the 
		imbalance ratios (numbers under the bars) increasing from left to right. 
	}
	\label{fig:accgain}
	\vspace{-0.4cm}
\end{figure}

\begin{figure}
	\centering
	\includegraphics[width=0.95\linewidth]{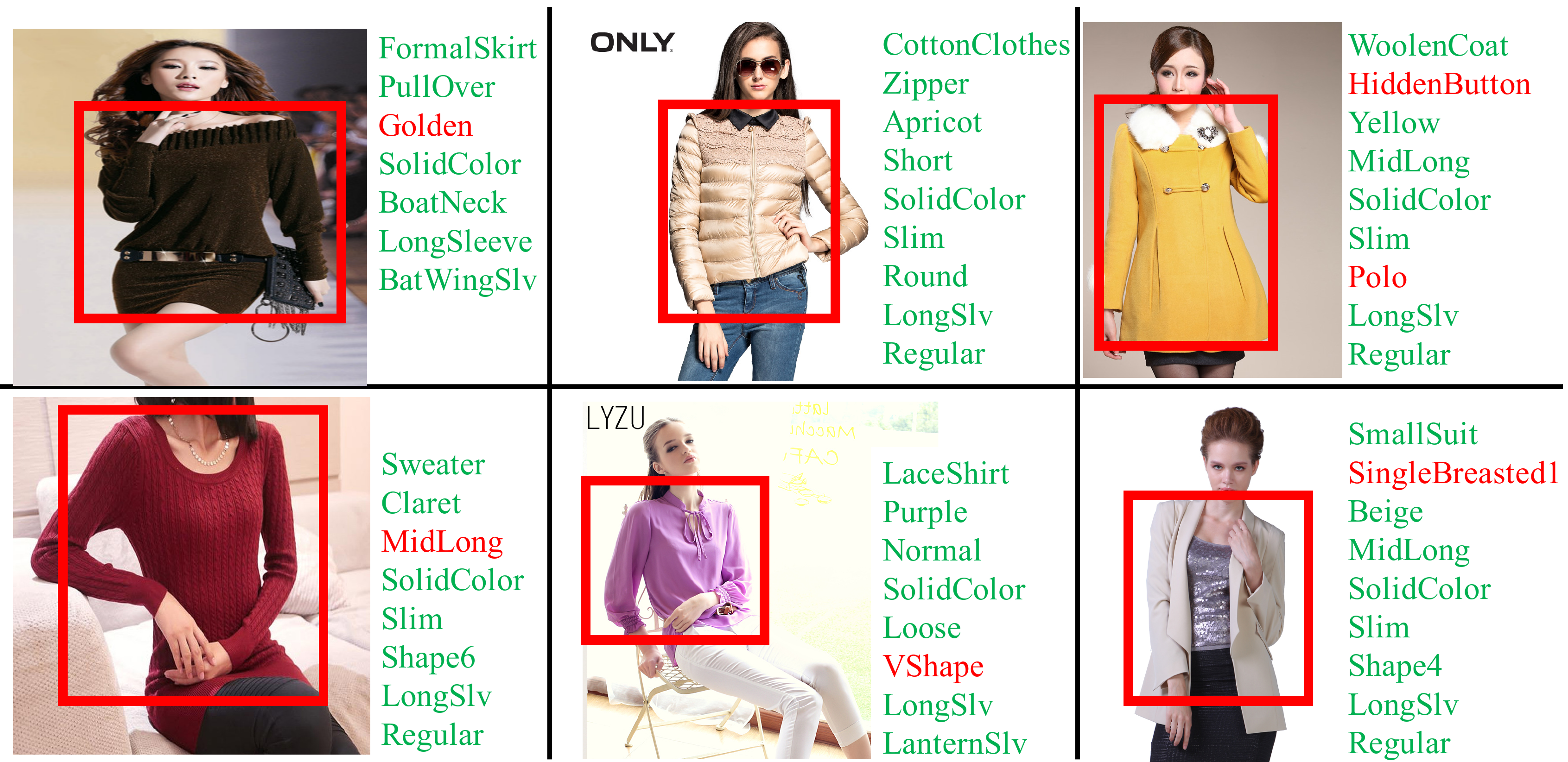}
	\vskip -0.3cm
	\caption{
		Examples of clothing attribute recognition by the CRL model,
		with false attribute prediction in red (Red box: clothing auto-detection).
	}
	\label{fig:vis_cloth}
	\vspace{-0.4cm}
\end{figure}

\vspace{0.1cm}
\noindent {\bf Further Analysis }
\sgg{We further examined the performance of CRL and LMLE
in comparison to the base model MTCT. 
Similar to CelebA, we split the 9 attributes into 
two groups at a 1:5 class imbalance ratio:
bottom-1 (the first column in Table \ref{tab:arts_clothing}) and top-8
(the remaining).
Figure \ref{fig:accgain} and Table \ref{tab:LMLEvsCRL_clothing}
show that:
{\bf(1)} LMLE improves MTCT on all clothing attributes
with varying imbalance ratios.
This suggests that 
LMLE does address the imbalanced data learning problem
in a multi-class setting
by embedding local class structures 
into deep feature learning.
{\bf (2)} Compared to LMLE, CRL achieves more significant performance gains on more severely imbalanced attributes.
On the top-8 imbalanced attributes, CRL achieves mean accuracy gain
of 7.10\% {\em versus} 2.10\% by LMLE
(Table \ref{tab:LMLEvsCRL_clothing}).}
In particular, our CRL improves LMLE by 10.03\% in accuracy
for recognising ``Sleeve Shape'',
a fine-grained and visually ambiguous attribute
due to its locality and subtle inter-class discrepancy
(Fig. \ref{fig:vis_cloth}).
This evidence is interesting as it shows that
class training data distribution affects a model's ability to learn
effectively fine-grained class discrimination. 
Importantly, a model's ability in coping effectively with class imbalanced
data learning can help improve its learning of fine-grained class discrimination.
This further demonstrates the strength of CRL over existing
models for mitigating model learning bias given severely imbalanced fine-grain
labelled classes in an end-to-end deep learning framework.
%
%
%
\\

\noindent {\bf Model Training Cost Analysis }
We examined the model training cost of LMLE and CRL 
on X-Domain 
using the same workstation as on CelebA.
We used the original author released codes with the suggested optimisation setting, e.g.
trained the LMLE for 4 rounds each with 5,000 CNN training iterations.
We started with the ImageNet-1K trained VGGNet16 features \cite{simonyan2014very}. 
For model training, LMLE took 429.9 hours, with each round taking 107.5 hours 
including 
27.6 hours for ``clustering+quintuplet construction'' and 
79.9 hours for ``CNN model optimisation''.
In contrast, CRL took 60.4 hours, that is 7.1 (429.9/60.4) times faster than LMLE.

\begin{table}[h]
	\centering
	\setlength{\tabcolsep}{0.15cm}
	\caption{
		\sgg{Evaluation of CRL on clothing attribute recognition
                with the DeepFashion benchmark \cite{liu2016deepfashion}. 
		Metric: {\em Class-balanced accuracy} (\%).} 
	}
	\vskip -0.3cm
	\label{tab:deepfashion}
	\begin{tabular}{c||c|c|c|c|c||c}
		\hline
		CRL & Texture & Fabric & Shape & Part & Style& \multirow{2}{*}{\em Mean}\\ \cline{1-6}  
		ImbRatio (1:x)& 733&393&314&350&149\\\hline\hline
		\xmark &53.29 & 52.86 & 53.02 & 51.25 & 51.20&  52.20
		\\ \hline
		\cmark &\bf 55.37 &\bf 55.02 &\bf 55.22 &\bf 53.90 &\bf 53.75& \bf 54.56
		\\  \hline
	\end{tabular}
	\vspace{-0.3cm}
\end{table}

\vspace{0.3cm}
\noindent {\bf Further Evaluation }
We further evaluated the CRL on the DeepFashion clothing
attribute dataset \cite{liu2016deepfashion} with a controlled experiment.
We adopted a test setting that is consistent with all the other experiments:
(1) The standard multi-label classification 
setting {\em without} using the clothing landmark and category labels 
(used in \cite{liu2016deepfashion}).
(2) ResNet50 \cite{he2016deep} as the base network trained by the CE loss.
(3) Top-5 attribute predictions in a class-balanced
accuracy metric other than a class-biased metric as in \cite{liu2016deepfashion}.
We adopted the standard data split: 
209,222/40,000/40,000 images for model training/validation/test.
We trained the deep models from scratch 
with the learning rate as 0.01, 
the decay as 0.00004, 
the batch size as 64,
and the epoch as 141.
We focused on evaluating the additional effect of CRL on top of
the CE loss.
Table \ref{tab:deepfashion} shows that CRL yields a 2.36\% (54.56-52.20) boost in mean accuracy.

\subsection{Comparisons on Object Category Recognition}
\label{sec:exp_cifar100}


We evaluated the CRL on a popular {\em class balanced} single-label object category benchmark CIFAR-100 \cite{krizhevsky2009learning}.

\vspace{0.1cm}
\noindent {\bf Network Architecture }
We evaluated the CRL in three state-of-the-art CNN models:
(1) CifarNet \cite{krizhevsky2009learning},
(2) ResNet32 \cite{he2016deep},
and (3) DenseNet \cite{huang2016densely}.
Each CNN model was trained by the conventional CE loss (Eqn. \eqref{eq:loss}).
The purpose is to test their performance gains in single-label object classification
when incorporating the proposed CRL regularisation (Eqn. \eqref{eq:final_loss}).

\vspace{0.1cm}
\noindent {\bf Parameter Settings }
We trained each CNN from scratch 
with the learning rate at 0.1, 
the decay at 0.0005,
the momentum at 0.9,
the batch size at 256, and
the epoch at 200.
In the class {\em balanced} test,
we cannot directly deploy our loss formulation
(Eqn. \eqref{eq:final_loss}) 
as the imbalance measure $\Omega_\text{imb}\!=\!0$
hence eliminating the CRL.
Instead, we integrated our CRL with the CE loss 
using equal weight by setting $\alpha\!=\!0.5$
for all models.
For the class imbalanced cases,
we set $\eta\!=\!0.01/0.5/0.5$ for CifarNet/ResNet32/DenseNet,
respectively.

\vspace{0.1cm}
\noindent {\bf (I) Comparative Evaluation }
Table \ref{tab:cifar100} shows the single-label object classification accuracy.
Interestingly, it is found that
our CRL approach can consistently improve 
state-of-the-art CNN models,
e.g. increasing the accuracy of CifarNet/ResNet32/DenseNet 
by $3.6\%$/$1.2\%$/$0.8\%$, respectively.
This shows that the advantages of our batch-wise minority class
rectification method remain on class balanced cases.
The plausible reasons are:
(1) Whilst the global class distribution is balanced, 
random sampling of mini-batch adopted by common deep learning 
may introduce some imbalance in each iteration. 
Our per-batch balancing strategy hence has the chance to regularise inter-class
margin and benefit the overall model learning. 
(2) The CRL considers the optimisation of class-level structural separation,
which can provide a complementary benefit
to the CE loss that instead performs per-sample single-class optimisation.

\begin{table}
	\centering
	\setlength{\tabcolsep}{0.3cm}
	\caption{
		Object classification performance (\%) on CIFAR-100 \cite{krizhevsky2009learning}.
	}
	\vskip -0.3cm
	\label{tab:cifar100}
	\begin{tabular}{c|c||c|c||c|c}
		\hline
		CifarNet 
		& 56.5 
		& ResNet32 
		& 68.1 
		& DenseNet 
		& 74.0 \\ \hline
		+{\bf CRL} & \bf +3.6 
		& +{\bf CRL} & \bf +1.2 
		& +{\bf CRL} & \bf +0.8\\ 
		\hline
	\end{tabular}
	\vspace{-0.3cm}
\end{table}

\begin{figure}
	\centering
	\includegraphics[width=1\linewidth]{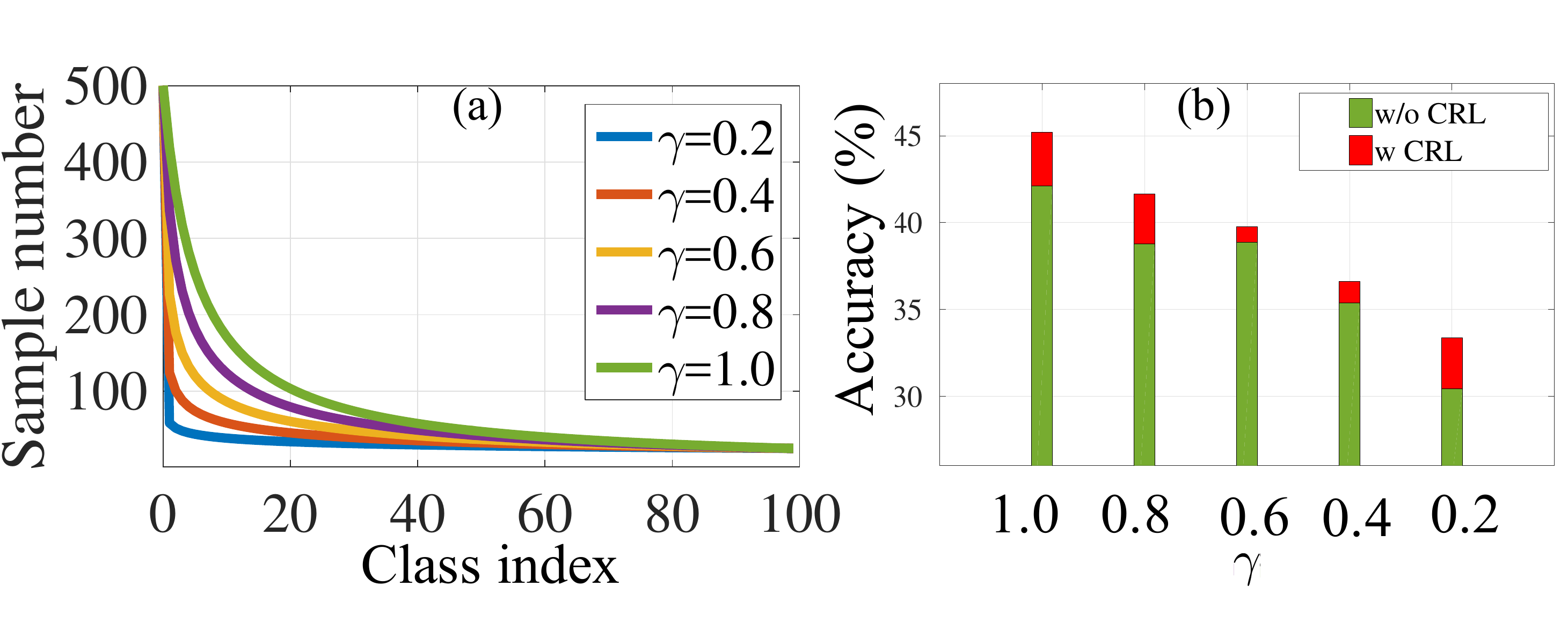}
	\vskip -0.5cm
	\caption{
	(a) Simulated imbalanced training data distributions on CIFAR-100 \cite{krizhevsky2009learning}.
	(b) Performance gains of ResNet32 by the CRL 
	on differently imbalanced training data.
	Metric: Mean {\em class-balanced accuracy} (\%).
	}
	\label{fig:imbalanceratio}
	\vspace{-0.3cm}
\end{figure}

\begin{table}
	\centering
	\setlength{\tabcolsep}{0.15cm}
	\caption{
		Effect of the CRL in different CNN models 
		given class imbalanced training data ($\gamma=1$).
		Metric: Mean {\em class-balanced accuracy} (\%).
	}
	\vskip -0.3cm
	\label{tab:ration}
	\begin{tabular}{c||c|c|c||c}
		\hline
		Training Dataset &CifarNet & ResNet32 & DenseNet & HOG+$k$NN \\ \hline \hline
		CIFAR-100$^\text{bln(1)}$ &38.6  &48.2  & 52.7 & 7.3 \\ \hline \hline
		CIFAR-100$^\text{imb(1)}$ &34.7 &42.1 &46.3 & 6.5 \\\hline
		& \multicolumn{4}{c}{\bf +CRL}\\\hline
		CIFAR-100$^\text{imb(1)}$ &\bf{36.7}  &\bf{45.2}  & \bf{49.5} & N/A  \\ \hline
	\end{tabular}
\vspace{-0.3cm}
\end{table}

\vspace{0.1cm}
\noindent {\bf (II) Effect Analysis of Imbalanced Training Data }
We further evaluated the deep model performance and
the CRL under different imbalance ratios. 
To this end, we carried out a controlled experiment
by simulating class imbalance cases in training data. 
Specifically: 
{\bf (1)} We simulated class imbalanced training data by 
a power-law class distribution as (Fig. \ref{fig:imbalanceratio}(a)):
$f_{\text{CS}}(i)=\frac{a}{i^{\gamma}+b}$, where
$i\! \in \!\{1,2,\cdots,100\}$ is the class index,
$\gamma$ represents a preset parameter for controlling the
imbalanced degree,
$a$ and $b$ are two numbers estimated by 
the largest ($500$) and smallest ($25$) class size. 
We call the resulted training set ``CIFAR-100$^{\text{imb}(\gamma)}$''.
{\bf (2)} \sgg{We constructed a corresponding 
dataset ``CIFAR-100$^{\text{bln}(\gamma)}$'', subject to
having the same number of images covering all classes as ``CIFAR-100$^{\text{imb}(\gamma)}$''
and being class balanced (i.e. all classes are equally sized)}.
%
This is necessary as
``CIFAR-100$^{\text{imb}(\gamma)}$'' and 
``CIFAR-100'' differ in both data balance and size
thus not directly comparable.
{\bf (3)} We trained the CNN models with and without CRL 
on these simulated training sets separately
and tested their performances on the same standard test data.
{\bf (4)} To compare deep learning methods with conventional models,
we also evaluated
the $k$-nearest neighbour classifier with the HOG feature \cite{dalal2005histograms}.
Table \ref{tab:ration} shows the results when $\gamma\!=\!1$.
We observed that:
(1) Given class imbalanced training data, all three CNN models are adversely affected,
with accuracy decreased by $3.9\%$ (CifarNet), $6.1\%$ (ResNet32), and
$6.4\%$ (DenseNet) respectively.
Interestingly, the stronger CNNs suffer more performance degradation.
(2) CRL improves all three CNN models by $2.0\%\!\!\sim\!\!3.2\%$ in
accuracy, which show the effectiveness of CRL.
(3) 
\sgg{All three deep learning models are sensitive
to imbalanced training data with similar relative
performance drops as the conventional non-deep-learning
HOG+$k$NN model. 
This suggests that deep learning models are
not necessarily superior in tackling the class imbalanced learning challenge.}
%
%
Moreover, we evaluated the CRL with ResNet32
given different imbalance cases $\gamma$ ranging from $0.2$ to $1.0$.
Figure \ref{fig:imbalanceratio} (b) shows its performance gains across all these settings.
\sgg{We observed no clear trend between model performance and $\gamma$ since their
relationship is non-linear.
In particular, the model generalisation depends
not only on the class distribution but also on other factors 
such as the specific training samples,
i.e. information content is variable (and unknown) over training samples.  
}

\subsection{Further Evaluations and Discussions}	
\label{sec:further_eval}
{We conducted component analysis
for providing more insights on CRL.
By default, we adopted the class-level relative comparison
based CRL (Eqn. ~\eqref{eq:bt}) 
and used the most imbalanced X-Domain dataset,
unless declared otherwise.}

\begin{table} [h]
	\centering
	\setlength{\tabcolsep}{0.28cm}
	\caption{Comparing hard mining schemes (Class/Instance) and 
		CRL loss functions
		(Relative ({\bf Rel}), Absolute ({\bf Abs}), and Distribution ({\bf Dis})).
		{Metric}: Gain in the mean {\em class-balanced accuracy} 
		(\%).
	}
	\vskip -0.3cm
	\label{tab:crl_comb}
	\begin{tabular}{c||c|c|c||c|c|c}
		\hline
		\multirow{1}{*}{Dataset} & \multicolumn{3}{c||}{CelebA \cite{liu2015deep}} & \multicolumn{3}{c}{X-Domain \cite{chen2015deep}} \\\hline \cline{1-7} 
		Loss Design & Abs & Rel & Dis & Abs & Rel & Dis \\ \hline \hline
		Instance Level & 5.30 & 5.45 & 3.45 & 4.90 &  6.46     & 2.87  \\ \hline
		Class Level    & 5.08 & {\bf6.32} & 4.90 &4.76  &{\bf 6.89}  & 4.32  \\ \hline
	\end{tabular}
	\vspace{-0.1cm}
\end{table}

\noindent {\bf CRL Design }
We evaluated the two hard mining schemes 
({\em class-level} and {\em instance-level}, Sec.~\ref{sec:method_hard_mining}), 
and 
three loss types 
({\em relative, absolute}, and {\em distribution comparison}, Sec. \ref{sec:method_CRL_function}).
We tested therefore 6 CRL design combinations
in the comparison with
baseline models without imbalanced learning: 
``DeepID2'' on CelebA and ``MTCT'' on X-domain.
%
%
%
%
%
%
%
%
Table \ref{tab:crl_comb} shows that:
(1) All CRL models improve the mean accuracy consistently,
with the CRL(Class+Rel) the best.
%
%
%
(2) With the same loss type, 
the class-level design is superior in most cases. 
This suggests that regularising the score space 
is more effective than 
the feature space. 
\sgg{A plausible explanation is that the former
is more compatible with the conventional CE loss
which also operates with class scores.}

\begin{figure}[h] 
	\centering
	\includegraphics[width=1\linewidth]{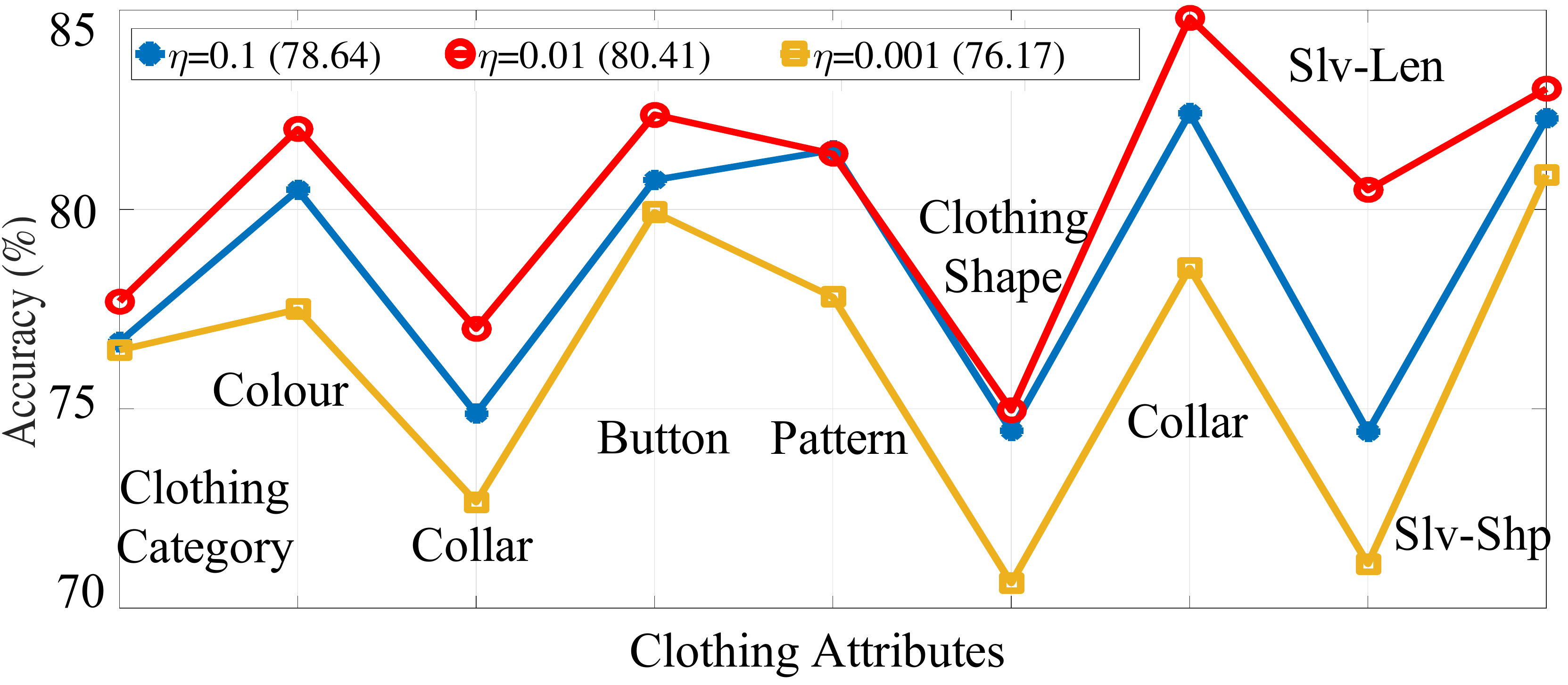}
	\vskip -0.3cm
	\caption{
		\sgg{Effect of the weight between the CE loss and our CRL on X-domain
		by varying $\eta$ in Eqn. \eqref{eq:final_loss}. 
		Metric: {\em Class-balanced accuracy} (\%).
		The mean accuracy for each setting is given in the parentheses.} 
	}
	\vspace{-0.2cm}
	\label{fig:loss_weight}
\end{figure}

\vspace{0.1cm}
\sgg{
	\noindent {\bf Loss Weight Optimisation }
	We evaluated the effectiveness of the weight between the CE loss
        and the CRL loss
	by tuning the coefficient $\eta$ in Eqn.~\eqref{eq:final_loss}
	on a range from $0.001$ to $0.1$ using the X-Domain benchmark.
	Figure \ref{fig:loss_weight} shows that
	the best weight selection is $\eta \!=\!0.01$.
	Moreover, it is found that the change in $\eta$ affects
	the performance on most or all attributes 
	consistently.
	This indicates that the CRL formulation with loss weighting is imbalance adaptive,
	capable of effectively modelling multiple attribute labels with diverse class imbalance ratios
	by a single-value hyper-parameter ($\eta$) optimisation
        using cross-validation.
	
	%
}

\begin{table}[th]
	\footnotesize
	\setlength{\tabcolsep}{0.15cm}
	\centering
	\caption{Effect of the CRL hard mining (HM) and joint
		learning (JL) in comparison to the LMLE on X-Domain. 
	} 
	\label{tab:hard_mining}
	\vskip -0.3cm
	\begin{tabular}{c||c|c|c}
		\hline
		Method & CRL(HM+JL) & CRL(JL) & LMLE \cite{huang2016learning} \\ \hline\hline
		Mean Accuracy (\%) 
		&{\bf80.42} 
		& 78.89 
		& 75.77 \\
		\hline
	\end{tabular}
	\vspace{-0.1cm}
%
%
\end{table}

\vspace{0.1cm}
\sgg{\noindent {\bf Hard Mining and Joint Learning}
	We further evaluated the individual effects of Hard Mining (HM) and
        Joint Learning (JL) the features and classifier in the CRL
        model ``CRL(HM+JL)'', as in comparison to LMLE\footnote{\sgg{In this evaluation,
	we treat LMLE as a whole without separating/removing its
        built-in hard mining mechanism.}} \cite{huang2016learning}. 
	Table \ref{tab:hard_mining} shows the 
	performance benefit of the proposed CRL Hard Mining (HM) as compared
	to CRL without HM ``CRL(JL)'', with a
	1.53\% (80.42-78.89) mean accuracy advantage on X-Domain. 
	It also shows that the joint learning in CRL has a mean accuracy
        advantage of 3.12\% (78.89-75.77) over LMLE which has no joint
        learning but has hard mining.
%
	%
	%
}

\begin{figure}[h] 
	\centering
	\includegraphics[width=0.8\linewidth]{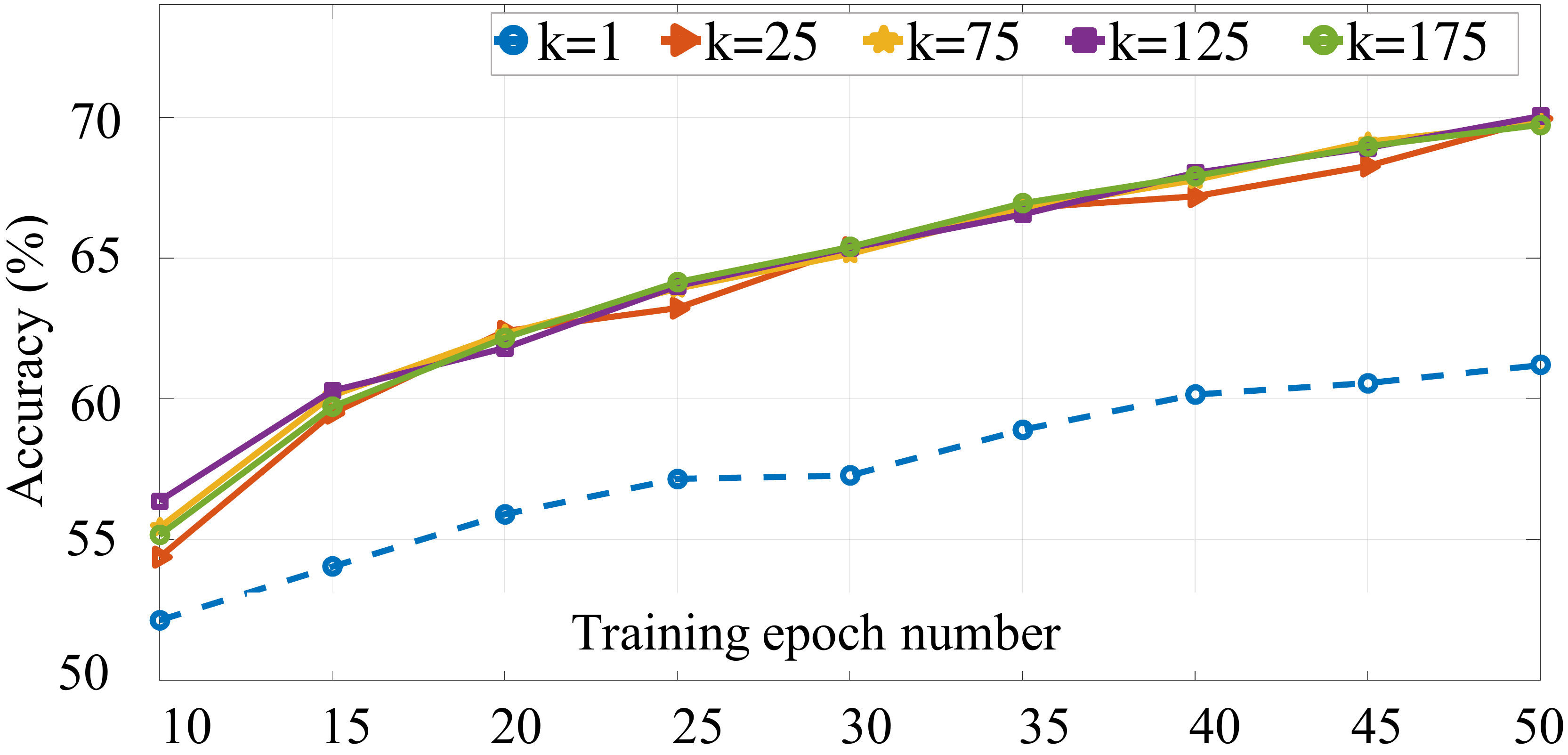}
	\vskip -0.2cm
	\caption{
		Effect of $\kappa$ (sample numbers) 
		in hard mining on X-Domain.
	}
	\vspace{-0.1cm}
	\label{fig:topkcurve}
\end{figure}

\vspace{0.1cm}
\noindent {\bf Top-$\kappa$ } 
We examined the effect of different
$\kappa$ values in hard mining from 1 to 175 
with step-size 25. 
Figure
\ref{fig:topkcurve} shows that when $\kappa\!\!=\!\!1$ 
(i.e. hardest mining), the model fails to capture a good converging trajectory.
This is because the hardest
mining represents over sparse and possibly incorrect (due to outlier noise) class boundaries, 
which hence causes poorer optimisation. 
When $\kappa\!\!>=\!\!25$, there is no
further improvement to model learning. Given that larger
$\kappa$ increases the model training cost, 
we set $\kappa\!=\!25$ for all our experiments.

\begin{table} [h] 
	\footnotesize
	\centering 
	\setlength{\tabcolsep}{0.3cm}
	\caption{
		Effect of the CRL class scope on X-Domain.
	}
	\vskip -0.3cm
	\label{tab:cls_mining}
	\begin{tabular}{c||c|c}
		\hline
		CRL Class Scope
		& Mean Accuracy (\%) & Training Time
		\\ \hline \hline
		
		Minority Classes 
		&{80.42}  &{\bf 60.4} Hours
		\\ 
		\hline 
		All Classes 
		&  {\bf 81.30} & {77.6} Hours\\
		\hline
	\end{tabular}
	\vspace{-0.1cm}
\end{table}

\vspace{0.1cm}
\noindent {\bf Class Scope }
We also evaluated the effect of the class scope (Minority-Classes and All-Classes) 
on which the CRL regularisation
is enforced in terms of both accuracy and model training cost.
Table \ref{tab:cls_mining} shows that applying the CRL to 
all classes in each batch yields superior performance.
\sgg{Specifically, relative to the baseline MTCT's $73.53\%$ mean accuracy 
	(Table \ref{tab:arts_clothing}),
	the scopes of Minority-Classes and All-Classes bring $6.89\%$ (80.42-73.53)
	and $7.78\%$ (81.30-73.53) accuracy gain, respectively.
	That is, the latter yields additional $12.9\%$ ((7.78-6.89)/6.89) gain 
	but at $28.5\%$ ((77.6-60.4)/60.4) extra training cost.}
%
%
This suggests better cost-effectiveness by focusing {\em only} on minority
classes in imbalanced data learning.

\begin{table}[h]
	\centering
	\caption{Effect of minority class criterion (Eqn. \eqref{eqn:minority_cls_definition}) on X-Domain.}
	\label{tab:minority_cls_definition}
	\vskip -0.3cm
	\begin{tabular}{c||c|c|c}
		\hline
		Minority Class Criterion & $\rho$=10\% & $\rho$=30\% & $\rho$=50\% ({\bf Ours}) \\\hline\hline
		Mean Accuracy (\%) & 77.82 & 79.29 & \bf 80.42 \\
		\hline 
	\end{tabular}
	\vspace{-0.1cm}
\end{table}

\vspace{0.1cm}
\noindent {\bf Minority Class Criterion }
At last, we evaluated the effect of minority class criterion 
($\rho$ in Eqn. \eqref{eqn:minority_cls_definition})
on a range from $10\%$ to $50\%$ 
to generalise the two-class minority class definition to a multi-class setting.
Table \ref{tab:minority_cls_definition} shows 
the effect on model performance when $\rho$ changes,
demonstrating that a minority class criterion setting with
$\rho\!=\!50\%$ is both most effective and conceptually
consistent with the two-class setting.

\section{Conclusion}
\label{conclusion}
In this work, we introduced an end-to-end class imbalanced deep
learning framework for large scale visual data learning.
%
%
The proposed Class Rectification Loss (CRL) approach is characterised by batch-wise
incremental minority class rectification with a scalable hard mining principle. 
Specifically, the CRL is designed to regularise the inherently biased deep model learning behaviour 
given extremely imbalanced training data. 
%
%
Importantly, CRL preserves the model optimisation convergence
characteristics of stochastic gradient descent,
therefore allowing for
efficient end-to-end deep learning on significantly imbalanced
training data with multi-label semantic interpretations.
Comprehensive experiments were carried out to show the clear advantages and scalability 
of the CRL method over not only the state-of-the-art imbalanced data
learning models but also dedicated deep learning visual recognition methods.
For example, the CRL surpasses the
best alternative LMLE by $3\%$ on the CelebA facial attribute benchmark and 
$5\%$ on the extremely imbalanced X-Domain clothing attribute benchmark, 
whilst enjoying over
7$\times$ faster model training advantage.
Our experiments also show the benefits of the CRL in 
learning standard deep models given class balanced training data.
%
Finally, we provided detailed component analysis for giving
insights into the characteristics of the CRL model design.

 
\section*{Acknowledgements}
{\small
We shall thank Victor Lempitsky for providing the histogram loss code,
Chen Huang and Chen Change Loy for sharing the pre-trained DeepID2 face recognition model. 
This work was partly supported by the China Scholarship Council, Vision Semantics Ltd., 
the Royal Society Newton Advanced Fellowship Programme (NA150459), 
and Innovate UK Industrial Challenge Project on Developing and Commercialising Intelligent Video Analytics Solutions for Public Safety (98111-571149).}

\ifCLASSOPTIONcaptionsoff
\newpage
\fi



\bibliographystyle{IEEEtran}
\bibliography{clothingAttr}

\begin{thebibliography}{10}
\providecommand{\url}[1]{#1}
\csname url@samestyle\endcsname
\providecommand{\newblock}{\relax}
\providecommand{\bibinfo}[2]{#2}
\providecommand{\BIBentrySTDinterwordspacing}{\spaceskip=0pt\relax}
\providecommand{\BIBentryALTinterwordstretchfactor}{4}
\providecommand{\BIBentryALTinterwordspacing}{\spaceskip=\fontdimen2\font plus
\BIBentryALTinterwordstretchfactor\fontdimen3\font minus
  \fontdimen4\font\relax}
\providecommand{\BIBforeignlanguage}[2]{{%
\expandafter\ifx\csname l@#1\endcsname\relax
\typeout{** WARNING: IEEEtran.bst: No hyphenation pattern has been}%
\typeout{** loaded for the language `#1'. Using the pattern for}%
\typeout{** the default language instead.}%
\else
\language=\csname l@#1\endcsname
\fi
#2}}
\providecommand{\BIBdecl}{\relax}
\BIBdecl

\bibitem{japkowicz2002class}
N.~Japkowicz and S.~Stephen, ``The class imbalance problem: A systematic
  study,'' \emph{Intelligent Data Analysis}, vol.~6, no.~5, pp. 429--449, 2002.

\bibitem{weiss2004mining}
G.~M. Weiss, ``Mining with rarity: a unifying framework,'' \emph{ACM SIGKDD
  Explorations Newsletter}, vol.~6, no.~1, pp. 7--19, 2004.

\bibitem{he2009learning}
H.~He and E.~A. Garcia, ``Learning from imbalanced data,'' \emph{IEEE TKDE},
  vol.~21, no.~9, pp. 1263--1284, 2009.

\bibitem{hospedales2013finding}
T.~M. Hospedales, S.~Gong, and T.~Xiang, ``Finding rare classes: Active
  learning with generative and discriminative models,'' \emph{IEEE TKDE},
  vol.~25, no.~2, pp. 374--386, 2013.

\bibitem{drummond2003c4}
C.~Drummond, R.~C. Holte \emph{et~al.}, ``C4.5, class imbalance, and cost
  sensitivity: why under-sampling beats over-sampling,'' in \emph{ICML
  Workshop}, 2003.

\bibitem{chawla2002smote}
N.~V. Chawla, K.~W. Bowyer, L.~O. Hall, and W.~P. Kegelmeyer, ``Smote:
  synthetic minority over-sampling technique,'' \emph{JAIR}, vol.~16, pp.
  321--357, 2002.

\bibitem{maciejewski2011local}
T.~Maciejewski and J.~Stefanowski, ``Local neighbourhood extension of smote for
  mining imbalanced data,'' in \emph{ICDM}, 2011.

\bibitem{ting2000comparative}
K.~M. Ting, ``A comparative study of cost-sensitive boosting algorithms,'' in
  \emph{ICML}, 2000.

\bibitem{tang2009svms}
Y.~Tang, Y.-Q. Zhang, N.~V. Chawla, and S.~Krasser, ``Svms modeling for highly
  imbalanced classification,'' \emph{IEEE TSMCB}, vol.~39, no.~1, pp. 281--288,
  2009.

\bibitem{Akbani-ecml04}
R.~Akbani, S.~Kwek, and N.~Japkowicz, ``Applying support vector machines to
  imbalanced datasets,'' in \emph{ECML}, 2004.

\bibitem{chen2015deep}
Q.~Chen, J.~Huang, R.~Feris, L.~M. Brown, J.~Dong, and S.~Yan, ``Deep domain
  adaptation for describing people based on fine-grained clothing attributes,''
  in \emph{CVPR}, 2015.

\bibitem{liu2015deep}
Z.~Liu, P.~Luo, X.~Wang, and X.~Tang, ``Deep learning face attributes in the
  wild,'' in \emph{ICCV}, 2015.

\bibitem{krawczyk2016learning}
B.~Krawczyk, ``Learning from imbalanced data: open challenges and future
  directions,'' \emph{Progress in Artificial Intelligence}, vol.~5, no.~4, pp.
  221--232, 2016.

\bibitem{gong2014person}
S.~Gong, M.~Cristani, S.~Yan, and C.~C. Loy, \emph{Person
  re-identification}.\hskip 1em plus 0.5em minus 0.4em\relax Springer, 2014,
  vol.~1.

\bibitem{feris2014attribute}
R.~Feris, R.~Bobbitt, L.~Brown, and S.~Pankanti, ``Attribute-based people
  search: Lessons learnt from a practical surveillance system,'' in
  \emph{ICMR}, 2014.

\bibitem{ChenEtAlcvpr13}
K.~Chen, S.~Gong, T.~Xiang, and C.~Loy, ``Cumulative attribute space for age
  and crowd density estimation,'' in \emph{CVPR}, 2013.

\bibitem{huang2016learning}
C.~Huang, Y.~Li, C.~Change~Loy, and X.~Tang, ``Learning deep representation for
  imbalanced classification,'' in \emph{CVPR}, 2016.

\bibitem{zhang2014panda}
N.~Zhang, M.~Paluri, M.~Ranzato, T.~Darrell, and L.~Bourdev, ``Panda: Pose
  aligned networks for deep attribute modeling,'' in \emph{CVPR}, 2014, pp.
  1637--1644.

\bibitem{triguero2015rosefw}
I.~Triguero, S.~del R{\'\i}o, V.~L{\'o}pez, J.~Bacardit, J.~M. Ben{\'\i}tez,
  and F.~Herrera, ``Rosefw-rf: the winner algorithm for the ecbdl’14 big data
  competition: an extremely imbalanced big data bioinformatics problem,''
  \emph{Knowledge-Based Systems}, vol.~87, pp. 69--79, 2015.

\bibitem{simonyan2014very}
K.~Simonyan and A.~Zisserman, ``Very deep convolutional networks for
  large-scale image recognition,'' 2015.

\bibitem{sharif2014cnn}
A.~Sharif~Razavian, H.~Azizpour, J.~Sullivan, and S.~Carlsson, ``Cnn features
  off-the-shelf: an astounding baseline for recognition,'' in \emph{CVPR},
  2014, pp. 806--813.

\bibitem{krizhevsky2012imagenet}
A.~Krizhevsky, I.~Sutskever, and G.~E. Hinton, ``Imagenet classification with
  deep convolutional neural networks,'' in \emph{NIPS}, 2012, pp. 1097--1105.

\bibitem{bengio2013representation}
Y.~Bengio, A.~Courville, and P.~Vincent, ``Representation learning: A review
  and new perspectives,'' \emph{IEEE TPAMI}, vol.~35, no.~8, pp. 1798--1828,
  2013.

\bibitem{zhou2006training}
Z.-H. Zhou and X.-Y. Liu, ``Training cost-sensitive neural networks with
  methods addressing the class imbalance problem,'' \emph{IEEE TKDE}, vol.~18,
  no.~1, pp. 63--77, 2006.

\bibitem{jeatrakul2010classification}
P.~Jeatrakul, K.~W. Wong, and C.~C. Fung, ``Classification of imbalanced data
  by combining the complementary neural network and smote algorithm,'' in
  \emph{International Conference on Neural Information Processing}, 2010.

\bibitem{alejo2006improving}
R.~Alejo, V.~Garc{\'\i}a, J.~M. Sotoca, R.~A. Mollineda, and J.~S. S{\'a}nchez,
  ``Improving the classification accuracy of rbf and mlp neural networks
  trained with imbalanced samples,'' in \emph{International Conference on
  Intelligent Data Engineering and Automated Learning}, 2006.

\bibitem{khoshgoftaar2010supervised}
T.~M. Khoshgoftaar, J.~Van~Hulse, and A.~Napolitano, ``Supervised neural
  network modeling: an empirical investigation into learning from imbalanced
  data with labeling errors,'' \emph{IEEE TNN}, vol.~21, no.~5, pp. 813--830,
  2010.

\bibitem{mazurowski2008training}
M.~A. Mazurowski, P.~A. Habas, J.~M. Zurada, J.~Y. Lo, J.~A. Baker, and G.~D.
  Tourassi, ``Training neural network classifiers for medical decision making:
  The effects of imbalanced datasets on classification performance,''
  \emph{Neural Networks}, vol.~21, no.~2, pp. 427--436, 2008.

\bibitem{huang2015cross}
J.~Huang, R.~S. Feris, Q.~Chen, and S.~Yan, ``Cross-domain image retrieval with
  a dual attribute-aware ranking network,'' in \emph{ICCV}, 2015.

\bibitem{liu2016deepfashion}
Z.~Liu, P.~Luo, S.~Qiu, X.~Wang, and X.~Tang, ``Deepfashion: Powering robust
  clothes recognition and retrieval with rich annotations,'' in \emph{CVPR},
  2016.

\bibitem{lin2014microsoft}
T.-Y. Lin, M.~Maire, S.~Belongie, J.~Hays, P.~Perona, D.~Ramanan,
  P.~Doll{\'a}r, and C.~L. Zitnick, ``Microsoft coco: Common objects in
  context,'' in \emph{ECCV}, 2014.

\bibitem{russakovsky2015imagenet}
O.~Russakovsky, J.~Deng, H.~Su, J.~Krause, S.~Satheesh, S.~Ma, Z.~Huang,
  A.~Karpathy, A.~Khosla, M.~Bernstein \emph{et~al.}, ``Imagenet large scale
  visual recognition challenge,'' \emph{IJCV}, vol. 115, no.~3, pp. 211--252,
  2015.

\bibitem{everingham2015pascal}
M.~Everingham, S.~A. Eslami, L.~Van~Gool, C.~K. Williams, J.~Winn, and
  A.~Zisserman, ``The pascal visual object classes challenge: A
  retrospective,'' \emph{IJCV}, vol. 111, no.~1, pp. 98--136, 2015.

\bibitem{krizhevsky2009learning}
A.~Krizhevsky and G.~Hinton, ``Learning multiple layers of features from tiny
  images,'' 2009.

\bibitem{griffin2007caltech}
G.~Griffin, A.~Holub, and P.~Perona, ``Caltech-256 object category dataset,''
  2007.

\bibitem{he2016deep}
K.~He, X.~Zhang, S.~Ren, and J.~Sun, ``Deep residual learning for image
  recognition,'' in \emph{CVPR}, 2016.

\bibitem{dong2016multi}
Q.~Dong, S.~Gong, and X.~Zhu, ``Multi-task curriculum transfer deep learning of
  clothing attributes,'' in \emph{WACV}, 2017.

\bibitem{arthur2006slow}
D.~Arthur and S.~Vassilvitskii, ``How slow is the k-means method?'' in
  \emph{ACM Annual Symposium on Computational Geometry}, 2006, pp. 144--153.

\bibitem{schroff2015facenet}
F.~Schroff, D.~Kalenichenko, and J.~Philbin, ``Facenet: A unified embedding for
  face recognition and clustering,'' in \emph{CVPR}, 2015.

\bibitem{han2005borderline}
H.~Han, W.-Y. Wang, and B.-H. Mao, ``Borderline-smote: a new over-sampling
  method in imbalanced data sets learning,'' in \emph{International Conference
  on Intelligent Computing}, 2005.

\bibitem{oquab2014learning}
M.~Oquab, L.~Bottou, I.~Laptev, and J.~Sivic, ``Learning and transferring
  mid-level image representations using convolutional neural networks,'' in
  \emph{CVPR}, 2014.

\bibitem{japkowicz2000learning}
N.~Japkowicz \emph{et~al.}, ``Learning from imbalanced data sets: a comparison
  of various strategies,'' in \emph{AAAI Workshop}, 2000.

\bibitem{barandela2003strategies}
R.~Barandela, J.~S. S{\'a}nchez, V.~Garc{\i}a, and E.~Rangel, ``Strategies for
  learning in class imbalance problems,'' \emph{Pattern Recognit.}, vol.~36,
  no.~3, pp. 849--851, 2003.

\bibitem{chen2004using}
C.~Chen, A.~Liaw, and L.~Breiman, ``Using random forest to learn imbalanced
  data,'' \emph{University of California, Berkeley}, 2004.

\bibitem{lin2002support}
Y.~Lin, Y.~Lee, and G.~Wahba, ``Support vector machines for classification in
  nonstandard situations,'' \emph{Machine Learning}, vol.~46, no.~1, pp.
  191--202, 2002.

\bibitem{liu2000improving}
B.~Liu, Y.~Ma, and C.~K. Wong, ``Improving an association rule based
  classifier,'' in \emph{European Conference on Principles of Data Mining and
  Knowledge Discovery}, 2000.

\bibitem{zadrozny2001learning}
B.~Zadrozny and C.~Elkan, ``Learning and making decisions when costs and
  probabilities are both unknown,'' in \emph{SIGKDD}, 2001, pp. 204--213.

\bibitem{quinlan1991improved}
J.~R. Quinlan, ``Improved estimates for the accuracy of small disjuncts,''
  \emph{Machine Learning}, vol.~6, no.~1, pp. 93--98, 1991.

\bibitem{wu2005kba}
G.~Wu and E.~Y. Chang, ``Kba: Kernel boundary alignment considering imbalanced
  data distribution,'' \emph{IEEE TKDE}, vol.~17, no.~6, pp. 786--795, 2005.

\bibitem{zadrozny2003cost}
B.~Zadrozny, J.~Langford, and N.~Abe, ``Cost-sensitive learning by
  cost-proportionate example weighting,'' in \emph{ICDM}, 2003.

\bibitem{provost2000machine}
F.~Provost, ``Machine learning from imbalanced data sets 101,'' in \emph{AAAI},
  2000.

\bibitem{krawczyk2015cost}
B.~Krawczyk and M.~Wo{\'z}niak, ``Cost-sensitive neural network with roc-based
  moving threshold for imbalanced classification,'' in \emph{International
  Conference on Intelligent Data Engineering and Automated Learning}, 2015.

\bibitem{yu2016odoc}
H.~Yu, C.~Sun, X.~Yang, W.~Yang, J.~Shen, and Y.~Qi, ``Odoc-elm: Optimal
  decision outputs compensation-based extreme learning machine for classifying
  imbalanced data,'' \emph{Knowledge-Based Systems}, vol.~92, pp. 55--70, 2016.

\bibitem{chen2006decision}
J.~Chen, C.-A. Tsai, H.~Moon, H.~Ahn, J.~Young, and C.-H. Chen, ``Decision
  threshold adjustment in class prediction,'' \emph{SAR and QSAR in
  Environmental Research}, vol.~17, no.~3, pp. 337--352, 2006.

\bibitem{wozniak2013hybrid}
M.~Wozniak, \emph{Hybrid classifiers: methods of data, knowledge, and
  classifier combination}.\hskip 1em plus 0.5em minus 0.4em\relax Springer,
  2013, vol. 519.

\bibitem{wang2012applying}
S.~Wang, Z.~Li, W.~Chao, and Q.~Cao, ``Applying adaptive over-sampling
  technique based on data density and cost-sensitive svm to imbalanced
  learning,'' in \emph{IJCNN}, 2012.

\bibitem{wozniak2014survey}
M.~Wo{\'z}niak, M.~Gra{\~n}a, and E.~Corchado, ``A survey of multiple
  classifier systems as hybrid systems,'' \emph{Information Fusion}, vol.~16,
  pp. 3--17, 2014.

\bibitem{lan2010investigation}
J.~Lan, M.~Y. Hu, E.~Patuwo, and G.~P. Zhang, ``An investigation of neural
  network classifiers with unequal misclassification costs and group sizes,''
  \emph{Decision Support Systems}, vol.~48, no.~4, pp. 582--591, 2010.

\bibitem{huang2006evaluation}
Y.-M. Huang, C.-M. Hung, and H.~C. Jiau, ``Evaluation of neural networks and
  data mining methods on a credit assessment task for class imbalance
  problem,'' \emph{Nonlinear Analysis: Real World Applications}, vol.~7, no.~4,
  pp. 720--747, 2006.

\bibitem{fernandez2011dynamic}
F.~Fern{\'a}ndez-Navarro, C.~Herv{\'a}s-Mart{\'\i}nez, and P.~A. Guti{\'e}rrez,
  ``A dynamic over-sampling procedure based on sensitivity for multi-class
  problems,'' \emph{Pattern Recognit.}, vol.~44, no.~8, pp. 1821--1833, 2011.

\bibitem{castro2013novel}
C.~L. Castro and A.~P. Braga, ``Novel cost-sensitive approach to improve the
  multilayer perceptron performance on imbalanced data,'' \emph{IEEE TNNLS},
  vol.~24, no.~6, pp. 888--899, 2013.

\bibitem{khan2017cost}
S.~H. Khan, M.~Hayat, M.~Bennamoun, F.~A. Sohel, and R.~Togneri,
  ``Cost-sensitive learning of deep feature representations from imbalanced
  data,'' \emph{IEEE TNNLS}, 2017.

\bibitem{shen2015deepcontour}
W.~Shen, X.~Wang, Y.~Wang, X.~Bai, and Z.~Zhang, ``Deepcontour: A deep
  convolutional feature learned by positive-sharing loss for contour
  detection,'' in \emph{CVPR}, 2015.

\bibitem{wang2016training}
S.~Wang, W.~Liu, J.~Wu, L.~Cao, Q.~Meng, and P.~J. Kennedy, ``Training deep
  neural networks on imbalanced data sets,'' in \emph{IJCNN}, 2016.

\bibitem{guan2015deep}
S.~Guan, M.~Chen, H.-Y. Ha, S.-C. Chen, M.-L. Shyu, and C.~Zhang, ``Deep
  learning with mca-based instance selection and bootstrapping for imbalanced
  data classification,'' in \emph{CIC}, 2015.

\bibitem{yan2015deep}
Y.~Yan, M.~Chen, M.-L. Shyu, and S.-C. Chen, ``Deep learning for imbalanced
  multimedia data classification,'' in \emph{ISM}, 2015.

\bibitem{felzenszwalb2010object}
P.~F. Felzenszwalb, R.~B. Girshick, D.~McAllester, and D.~Ramanan, ``Object
  detection with discriminatively trained part-based models,'' \emph{IEEE
  TPAMI}, vol.~32, no.~9, pp. 1627--1645, 2010.

\bibitem{shrivastava2016training}
A.~Shrivastava, A.~Gupta, and R.~Girshick, ``Training region-based object
  detectors with online hard example mining,'' in \emph{CVPR}, 2016.

\bibitem{oh2016deep}
H.~Oh~Song, Y.~Xiang, S.~Jegelka, and S.~Savarese, ``Deep metric learning via
  lifted structured feature embedding,'' in \emph{CVPR}, 2016.

\bibitem{wang2015unsupervised}
X.~Wang and A.~Gupta, ``Unsupervised learning of visual representations using
  videos,'' in \emph{ICCV}, 2015.

\bibitem{Ustinova2016hist}
E.~Ustinova and V.~Lempitsky, ``Learning deep embeddings with histogram loss,''
  in \emph{NIPS}, 2016.

\bibitem{wang2014learning}
J.~Wang, Y.~Song, T.~Leung, C.~Rosenberg, J.~Wang, J.~Philbin, B.~Chen, and
  Y.~Wu, ``Learning fine-grained image similarity with deep ranking,'' in
  \emph{CVPR}, 2014.

\bibitem{liu2009learning}
T.-Y. Liu, ``Learning to rank for information retrieval,'' \emph{Foundations
  and Trends in Information Retrieval}, vol.~3, no.~3, pp. 225--331, 2009.

\bibitem{chopra2005learning}
S.~Chopra, R.~Hadsell, and Y.~LeCun, ``Learning a similarity metric
  discriminatively, with application to face verification,'' in \emph{CVPR},
  2005.

\bibitem{lecun1998gradient}
Y.~LeCun, L.~Bottou, Y.~Bengio, and P.~Haffner, ``Gradient-based learning
  applied to document recognition,'' \emph{Proceedings of the IEEE}, vol.~86,
  no.~11, pp. 2278--2324, 1998.

\bibitem{ioffe2015batch}
S.~Ioffe and C.~Szegedy, ``Batch normalization: Accelerating deep network
  training by reducing internal covariate shift,'' \emph{arXiv}, 2015.

\bibitem{sun2014deep}
Y.~Sun, Y.~Chen, X.~Wang, and X.~Tang, ``Deep learning face representation by
  joint identification-verification,'' in \emph{NIPS}, 2014.

\bibitem{evgeniou2004regularized}
T.~Evgeniou and M.~Pontil, ``Regularized multi--task learning,'' in
  \emph{SIGKDD}, 2004, pp. 109--117.

\bibitem{ando2005framework}
R.~K. Ando and T.~Zhang, ``A framework for learning predictive structures from
  multiple tasks and unlabeled data,'' \emph{JMLR}, vol.~6, no. Nov, pp.
  1817--1853, 2005.

\bibitem{vedaldi08vlfeat}
A.~Vedaldi and B.~Fulkerson, ``{VLFeat}: An open and portable library of
  computer vision algorithms,'' \url{http://www.vlfeat.org/}, 2008.

\bibitem{lin2013network}
M.~Lin, Q.~Chen, and S.~Yan, ``Network in network,'' \emph{arXiv}, 2013.

\bibitem{huang2016densely}
G.~Huang, Z.~Liu, K.~Q. Weinberger, and L.~van~der Maaten, ``Densely connected
  convolutional networks,'' in \emph{CVPR}, 2017.

\bibitem{dalal2005histograms}
N.~Dalal and B.~Triggs, ``Histograms of oriented gradients for human
  detection,'' in \emph{CVPR}, 2005.

\end{thebibliography}
%
%
%

%





\end{document}